\documentclass[10pt,twocolumn,letterpaper]{article}

\usepackage[pagenumbers]{wacv} 

\usepackage{graphicx}
\usepackage{amsmath}
\usepackage{amssymb}
\usepackage{booktabs}

\usepackage{float}

\usepackage{subfigure}

\usepackage{color}
\usepackage{multirow}

\usepackage[accsupp]{axessibility}

\usepackage[pagebackref,breaklinks,colorlinks]{hyperref}

\hypersetup{
    colorlinks=true,
    linkcolor=blue,
    filecolor=magenta,      
    urlcolor=blue,
    pdftitle={Overleaf Example},
    pdfpagemode=FullScreen,
    }

\usepackage[capitalize]{cleveref}
\crefname{section}{Sec.}{Secs.}
\Crefname{section}{Section}{Sections}
\Crefname{table}{Table}{Tables}
\crefname{table}{Tab.}{Tabs.}

\def\wacvPaperID{2199} 
\def\confName{WACV}
\def\confYear{2024}

\begin{document}

\title{\vspace{-0.2in}TPSeNCE: Towards Artifact-Free Realistic Rain Generation \\for Deraining and Object Detection in Rain}


\author{
Shen Zheng\textsuperscript{1}, Changjie Lu\textsuperscript{2}, Srinivasa G. Narasimhan\textsuperscript{1} \\
\textsuperscript{1}Carnegie Mellon University, \textsuperscript{2}University of Illinois Urbana-Champaign \\
{\tt\small \{shenzhen, srinivas\}@andrew.cmu.edu}, {\tt\small cl140@illinois.edu}
}

\maketitle

\begin{abstract}
  Rain generation algorithms have the potential to improve the generalization of deraining methods and scene understanding in rainy conditions. However, in practice, they produce artifacts and distortions and struggle to control the amount of rain generated due to a lack of proper constraints. In this paper, we propose an unpaired image-to-image translation framework for generating realistic rainy images. We first introduce a Triangular Probability Similarity (TPS) constraint to guide the generated images toward clear and rainy images in the discriminator manifold, thereby minimizing artifacts and distortions during rain generation. Unlike conventional contrastive learning approaches, which indiscriminately push negative samples away from the anchors, we propose a Semantic Noise Contrastive Estimation (SeNCE) strategy and reassess the pushing force of negative samples based on the semantic similarity between the clear and the rainy images and the feature similarity between the anchor and the negative samples. Experiments demonstrate  realistic rain generation with minimal artifacts and distortions, which benefits image deraining and object detection in rain. Furthermore, the method can be used to generate realistic snowy and night images, underscoring its potential for broader applicability. Code is available at \href{https://github.com/ShenZheng2000/TPSeNCE}{https://github.com/ShenZheng2000/TPSeNCE}. 
  
  
  \end{abstract}

\section{Introduction} 

\begin{figure}[t]
    \centering
    \includegraphics[width=\columnwidth]{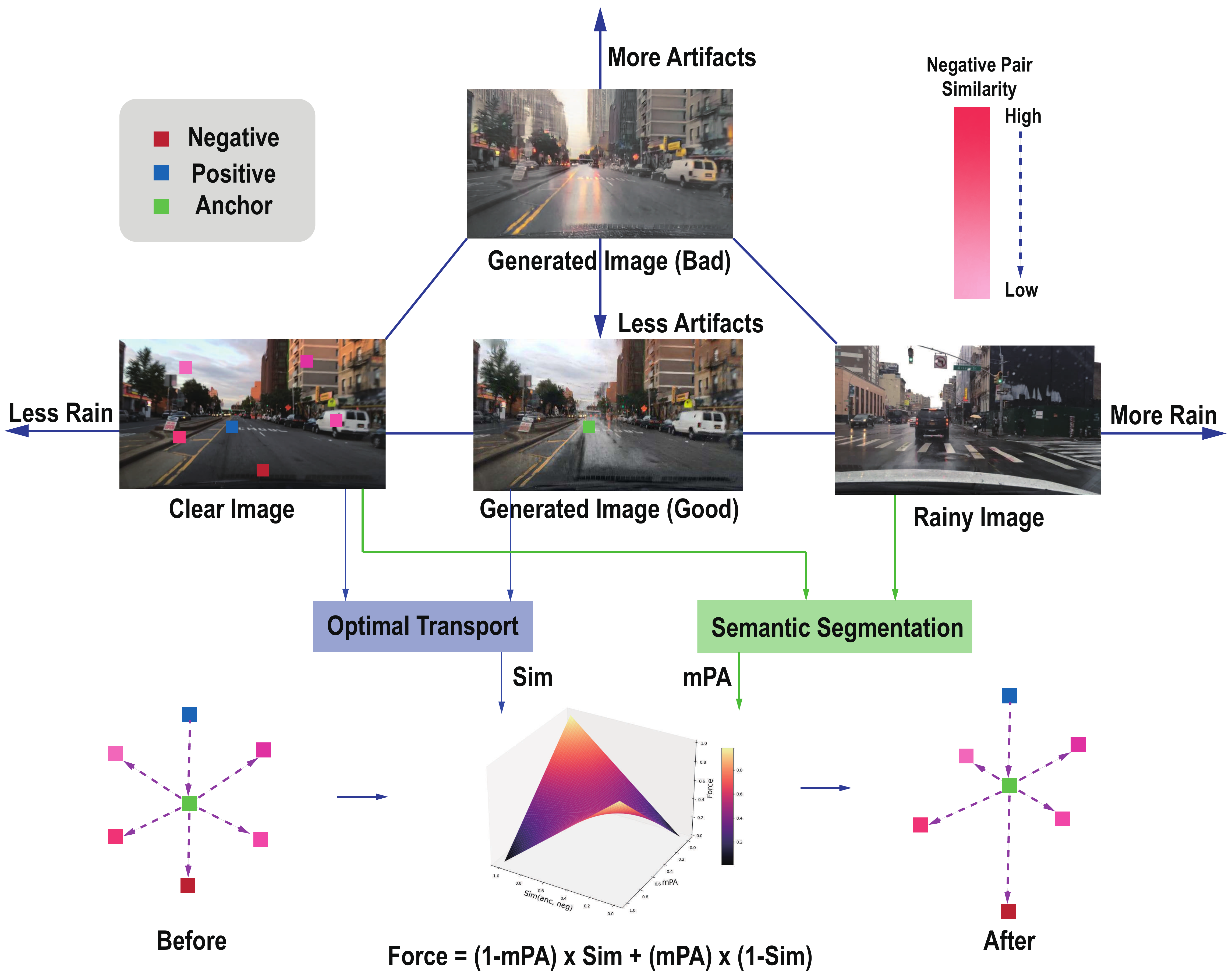}
    \caption{\textbf{Illustration for the proposed Triangular Probability Similarity (TPS) and Semantic Noise Contrastive Estimation (SeNCE) for rain generation}. TPS drives the generated rainy image towards the clear and the rainy image in the discriminator manifold to suppress artifacts and distortions. SeNCE collaboratively re-weights the pushing force of the negative patches based on their feature similarity with the anchor patch and refines that force with the semantic similarity between the clear and rainy image.}
    \label{fig:titlepage}
    \vspace{-0.1in}
\end{figure}



    Rain is a common bad weather condition that can significantly impair the quality of images and videos. Rain streaks, especially during heavy rain, obscure scene details and textures. Raindrops create a layer of water droplets on windshields, making objects appear blurry and distorted. Shiny wet roads create object reflections. Rain mist scatters ambient light, reducing the visibility of distant objects \cite{yang2020single}. These visual manifestations of rain not only impair the perceptual quality of images but also pose challenges for scene understanding algorithms like object detection, which are typically trained using data captured under clear weather conditions \cite{hnewa2020object}.

    One common way to improve object detection in the rain is to apply deraining (i.e. rain removal) as a preprocessing step.  Ideally, deraining algorithms should remove rain from images before applying object detection models. However, most state-of-the-art deraining methods \cite{fu2017removing, yang2017deep, ren2019progressive, wang2021rain, mehri2021mprnet, guo2021efficientderain, zheng2022sapnet} rely on supervised training with paired synthetic clear/rainy images due to the intractability of obtaining real paired clear/rainy images. Unfortunately, these methods do not generalize well to real-world rainy images \cite{yang2020single} because of the large domain gap between synthetic and natural rainy images. Although some deraining methods \cite{wei2019semi, yasarla2020syn2real, huang2021memory} use unpaired real clear/rainy images for unsupervised learning to improve generalization, it is challenging to integrate knowledge from the supervised and unsupervised branches seamlessly to enhance deraining performance on real data \cite{wang2019survey}.
    

    Another approach to enhance object detection in rainy conditions is using rain generation techniques to create synthetic rainy images for training object detectors. However, traditional model-based rain generation approaches such as those presented in \cite{rousseau2006realistic,garg2006photorealistic,mizukami2008realistic} rely on oversimplified assumptions and hand-crafted priors, which fail to accurately model the diverse types of real rain. In contrast, data-driven deep learning approaches such as unpaired image-to-image translation methods like UNIT \cite{liu2017unsupervised} have demonstrated their ability to translate images across different weather conditions. Nevertheless, these methods often produce artifacts and distortions while generating rain due to the lack of proper constraints. Additionally, controlling the amount of rain produced is challenging, as generating too much rain can lead to overlapping of the background and feature loss while generating too little rain results in an unrealistic-looking image. The presence of unwanted artifacts, distortions, and uncontrollable rain amounts can decrease the perceptual quality and hinder detection algorithms.
    

    In this paper, we address the above issues of rain generation approaches and propose an unpaired image-to-image translation framework for rain generation. Our analysis of the output matrix from the discriminator reveals a triangular relationship among the clear image, the generated rainy image,  and the real rainy image on the discriminator manifold (Fig. \ref{fig:TPS}). We observe that the generated rainy images with fewer artifacts and distortions are closer to the line segment connecting the clear and rainy images. Based on this observation, we propose a Triangular Probability Similarity (TPS) loss to guide the generated rainy images toward the real and clear images, thereby minimizing the artifacts and distortions. We then revisit the contrastive learning strategy of CUT \cite{park2020contrastive} and find that the amount of rain generated can be controlled by regulating the pushing force of contrastive learning. For this, we propose a Semantic Noise Contrastive Estimation strategy (SeNCE) that reweights the pushing force of negative pairs based on the similarity between the negatives and anchor, and the mean Pixel Accuracy (mPA) (shown in Fig. \ref{fig:SeNCE}) between the semantic segmentation maps of the clear and rainy images. 
    
    We evaluate the proposed method against multiple image-to-image translation approaches on various driving datasets including BDD100K \cite{yu2020bdd100k}, INIT \cite{shen2019towards}, and Boreas \cite{burnett2023boreas}. The evaluation on the BDD100K encompasses image-to-image translation, image deraining, and object detection. whereas the evaluation on INIT and Boreas focus solely on image-to-image translation.
    


    In summary, we present an unpaired image-to-image translation framework for generating realistic rainy images, with the following technical contributions:

    \begin{itemize}
        \item We introduce a Triangular Probability Similarity (TPS) loss to minimize the artifacts and distortions during rain generation. 
        \item We propose a Semantic Noise Contrastive Estimation (SeNCE) strategy to regularize the contrastive learning force to optimize the amounts of generated rain. 
        \item Our evaluation highlights the benefits of realistic rainy image generation for real rain removal and object detection in real rainy conditions. 
        
        
    \end{itemize}
    Beyond rain, our method can be used to generate realistic snowy images and night images, underscoring its potential for a broader applicability.


\begin{figure*}[t]
    \centering
    \includegraphics[width=\textwidth]{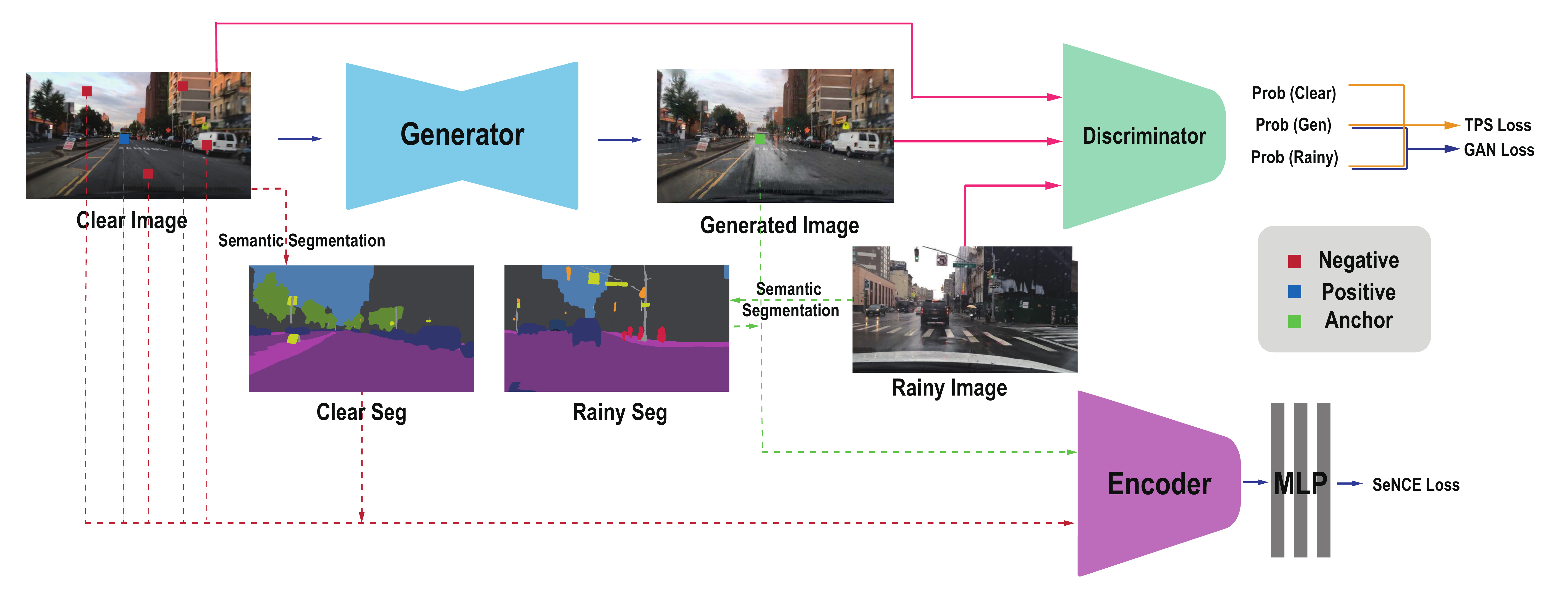}
    \caption{\textbf{Workflow for the proposed method}. The generator translates the clear image to generated (rainy) image. The discriminator receives the clear, generated, and rainy images to compute the TPS and GAN (adversarial) losses. Meanwhile, the encoder randomly selects and then embeds one positive patch and multiple negative patches from the clear image and the anchor patches from the generated image. The MLPs then process these embedded patches in a contrastive learning manner and output the SeNCE loss with the guidance of the semantic segmentation maps from the clear and rainy images. }
    \label{fig:workflow}
\end{figure*}

\section{Related Works}







    

    
    



\subsection{Unpaired Image-to-Image Translation}

    Unpaired image-to-image (I2I) methods aim to map between two domains of unpaired images. The main challenge is preserving source content while adopting the target style. CycleGAN \cite{zhu2017unpaired} introduced cycle-consistency loss to ensure content is kept by minimizing post-translation differences. UNIT \cite{liu2017unsupervised} built on this with the shared latent space assumption, suggesting that different domain images map to one latent code. However, these methods can hinder output style diversity. MUNIT \cite{huang2018multimodal} and DRIT \cite{lee2018diverse} then disentangle images into domain-invariant content and domain-specific style codes, facilitating multimodal I2I translation.

    Current I2I networks are designed for simple and common benchmarks, such as horse2zebra or apple2orange. When applied to more intricate tasks like rain generation, these general and weak constraints under unstable GAN training can misrepresent the intricate details, leading to unwanted artifacts and visual distortions.
    
    

\subsection{Contrastive Learning}

   Contrastive learning aims to attract positive instances towards an anchor while repelling negative ones. CUT \cite{park2020contrastive} was a pioneer to apply this to unpaired image-to-image translation by maximizing mutual information for random patches. However, CUT doesn't differentiate negative samples based on their similarity to the anchor. In response, QS-Attn \cite{hu2022qs} uses a query-selected attention module to pick critical negative patches. Meanwhile, NEGCUT \cite{wang2021instance} developed an instance-aware generator for hard negative examples, whereas MoNCE \cite{zhan2022modulated} employs an optimal-transport based approach to adjust the repulsion of negative pairs.
    
    While these methods advance general I2I tasks, they overlook the high-level semantic similarity between source and target images, instead focusing on the image-level feature similarity between the source and the generated image. In rain generation, lacking high-level cues from the target rainy image makes it hard to determine the appropriate rain amount for a realistic output.


\begin{figure}[t]
    \centering
    \subfigure[Clear (X)]{\includegraphics[width=0.24\columnwidth]{LaTeX/TPS/Clear.jpg}}
     \subfigure[Good (Z+)]{\includegraphics[width=0.24\columnwidth]{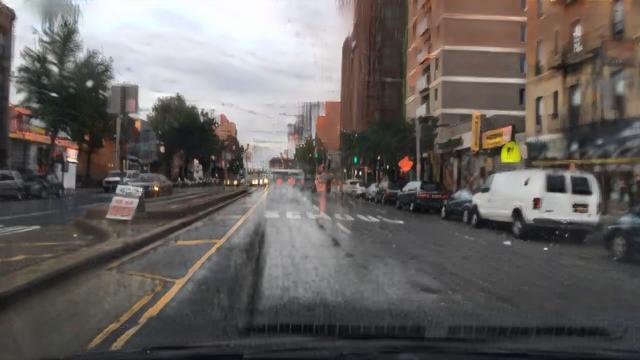}} 
    \subfigure[Bad (Z-)]{\includegraphics[width=0.24\columnwidth]{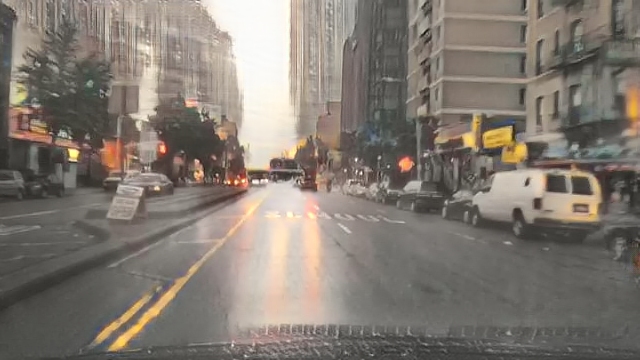}}
    \subfigure[Rainy (Y)]{\includegraphics[width=0.24\columnwidth]{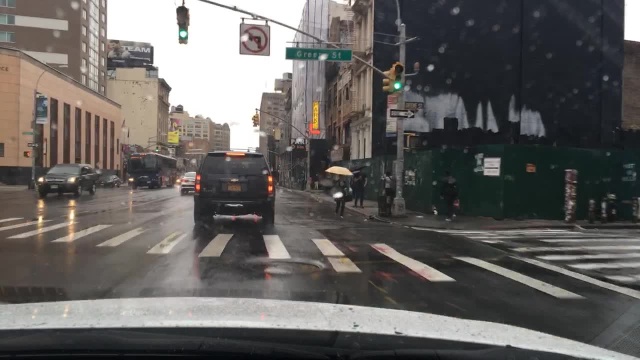}} \\

    \vspace{-0.3cm}

     \subfigure[T-SNE]{ \includegraphics[width=6cm]{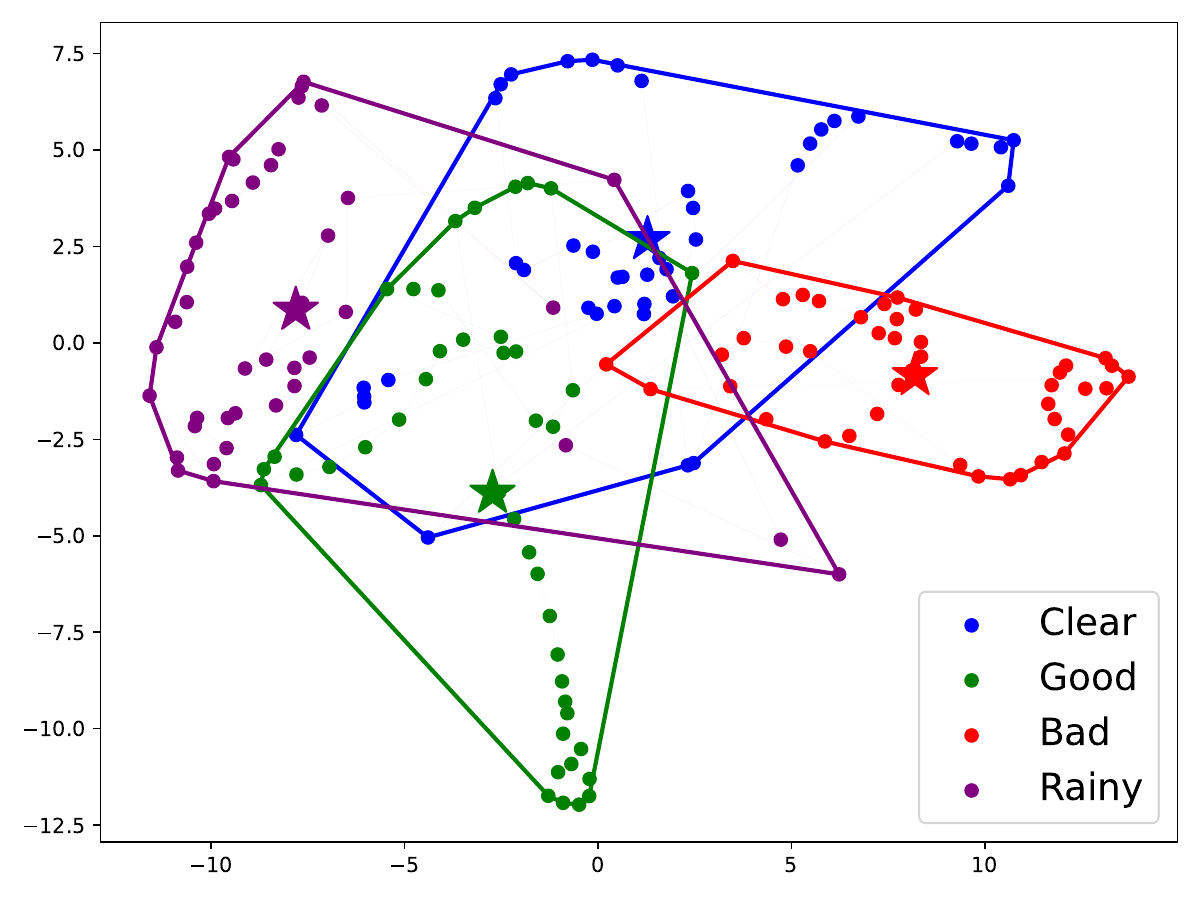} }
    \caption{\textbf{T-SNE \cite{van2008visualizing} visualization of the output matrices from the discriminator $D$}. Output matrices from different images are visualized with different colors. The centroid for each group is visualized as a star. Compared with $D(Z-)$ from the `Bad' generated image $Z-$ (i.e., images with more artifacts), $D(Z+)$ from the `Good' generated rainy image $Z+$ (i.e., images with fewer artifacts) have more area overlap with $D(X)$ and $D(Y)$. Designing a loss that brings $D(Z)$ near $D(X)$ and $D(Y)$ would benefit artifacts reduction, resulting in high-quality generated rainy images.  }
    \label{fig:TPS}
\end{figure}

\begin{figure}[t]
    \centering
    \includegraphics[width=\columnwidth]{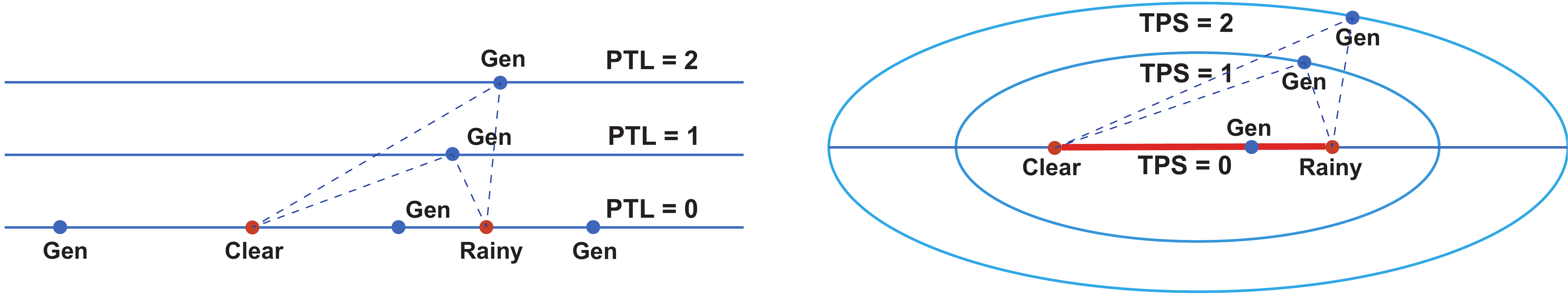}
    \caption{\textbf{Comparing Point To Line (PTL) with Triangular Probability Similarity (TPS)}: PTL employs unconstrained optimization, pushing the generated image `Gen' onto the straight line connecting `Clear' and `Rainy', frequently ends up on the extension cords. In contrast, TPS uses constrained optimization, adjusting the ellipse level set to position the generated image on the \color{red}{\textbf{line segment}} \color{black}{connecting `Clear' and `Rainy'. Zoom in for better view}.  }
    \label{fig:PTL_TPS}
\end{figure}

\section{Proposed Methods}


 In this section, we first explain the Triangular Probability Similarity (TPS) loss and revisit the Noise Contrastive Estimation (NCE) \cite{gutmann2010noise} schemes of CUT \cite{park2020contrastive} and MoNCE \cite{zhan2022modulated}. Extending these NCEs, we derive our Semantic Noise Contrastive Estimation (SeNCE) strategy. Last, we display the loss functions for model training. An overview of the proposed method's workflow is shown in Fig. \ref{fig:workflow}.
    



    \subsection{Triangular Probability Similarity (TPS)}



    
   Generating realistic rainy images while minimizing artifacts and distortions is a challenging task. Due to the ill-posed nature of rain generation and the instability of GAN training \cite{goodfellow2020generative}, the generated rainy images often suffer from artifacts and distortions.
    
    We show a T-SNE visualization in Fig. \ref{fig:TPS}, which explains the motivation for our Triangular Probability Similarity (TPS) loss.  Let $X$ be the clear image, $Y$ be the rainy image, and $Z$ be the generated rainy image. The TPS loss is based on the output representation from the discriminator $D$. It constrains $D(Z)$ to lie in the space spanned by $D(X)$ and $D(Y)$, ensuring that the generated rainy image follows a similar distribution as the clear image and the real rainy image. This strategy effectively mitigates undesired artifacts and distortions, as the information for the generated images is sourced exclusively from the clear image (providing background) and the real rainy image (providing rain).
    

    A potential issue with calculating TPS based on the distance between $D(Z)$ and the \textbf{straight line} connecting the centroid of $D(X)$ and $D(Y)$ is that it may guide $D(Z)$ towards the extension cord of either $D(X)$ or $D(Y)$, leading to \textit{too much or too little rain} in the generated image (as shown in Fig. \ref{fig:titlepage}). Additionally, $D(Z)$ may end up too far from $D(X)$ and $D(Y)$, resulting in \textit{artifacts and distortions} in the generated image that do not belong to either $X$ or $Y$.
    
    

    As shown in Fig. \ref{fig:PTL_TPS}, we address these issues with a loss function based on the distance from $D(Z)$ to the \textbf{line segment} between the centroid of $D(X)$ and $D(Y)$. Based on the triangular inequality, we use the following TPS loss:

    \begin{equation}
        \begin{aligned}
            \mathcal{L}_{\text{TPS}}(X, Y, Z)= \frac{1}{HW}
            &\sum_{i=1}^{H} \sum_{j=1}^{W} (|D(X)_{i,j} -  D(Z)_{i,j}|  \\
            &+ |D(Y)_{i,j} - D(Z)_{i,j}| \\
            &-|D(X)_{i,j} - D(Y)_{i,j}|)
        \end{aligned}
    \end{equation}

    Where $H$ and $W$ represent the height and width of the probability matrix $D(Z)$, respectively.

\begin{figure}[t]
    \centering
    \subfigure[Clear Img]{\includegraphics[width=0.24\columnwidth]{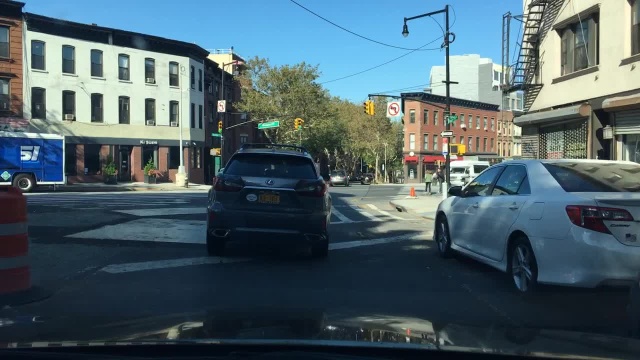}}
    \subfigure[Rainy Img 1]{\includegraphics[width=0.24\columnwidth]{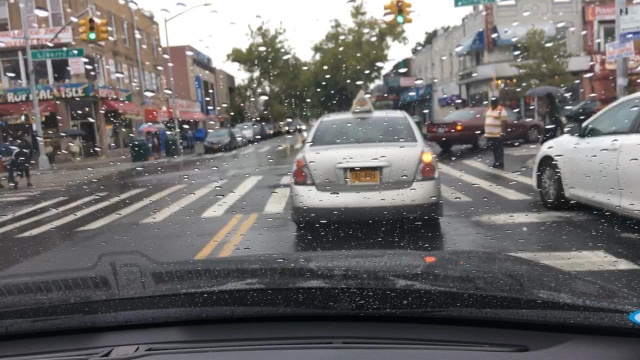}}
    \subfigure[Rainy Img 2]{\includegraphics[width=0.24\columnwidth]{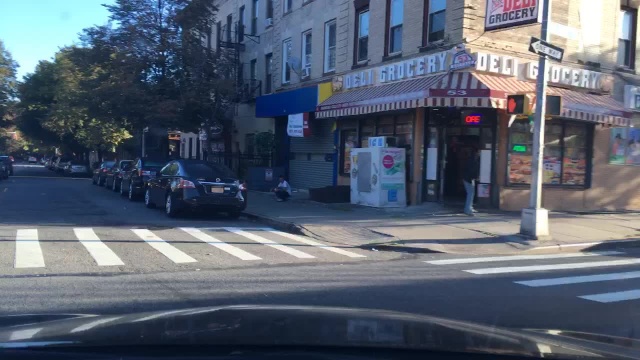}}
    \subfigure[Rainy Img 3]{\includegraphics[width=0.24\columnwidth]{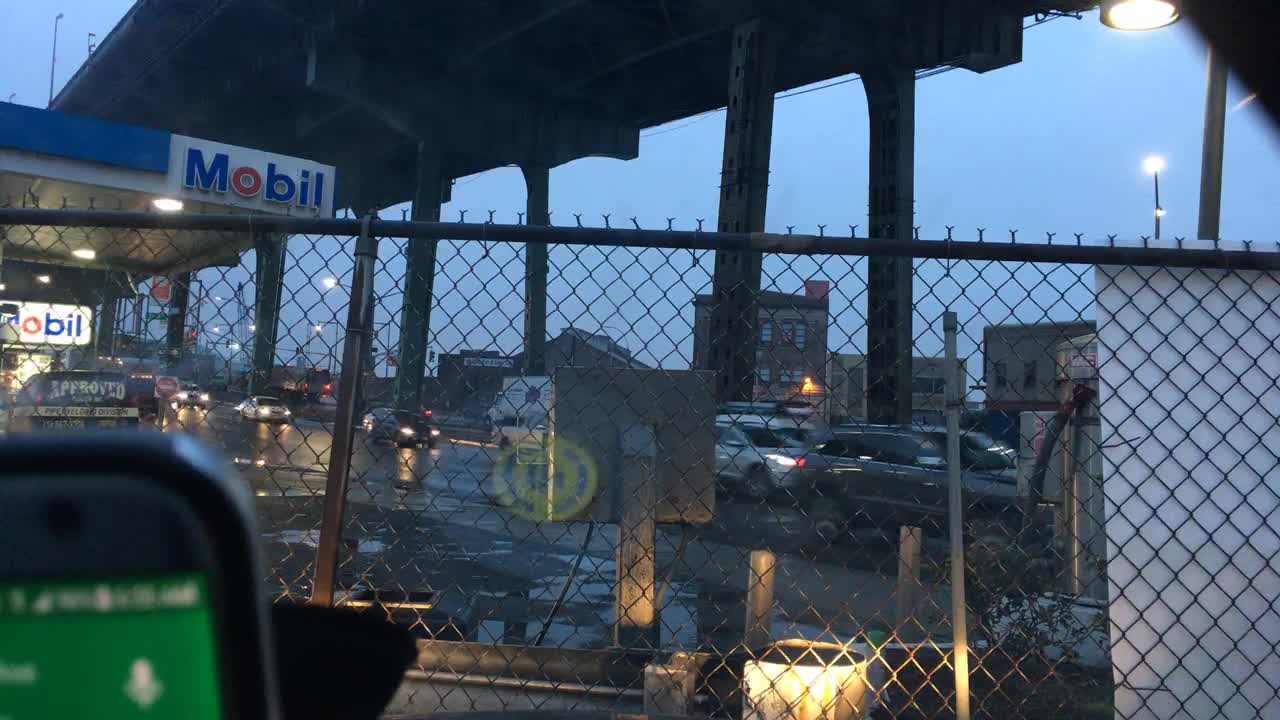}} \\

    \vspace{-0.3cm}

   \subfigure[Clear Seg]{\includegraphics[width=0.24\columnwidth]{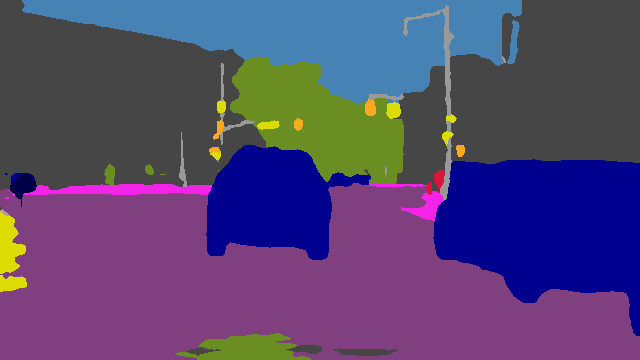}} 
    \subfigure[Rainy Seg 1]{\includegraphics[width=0.24\columnwidth]{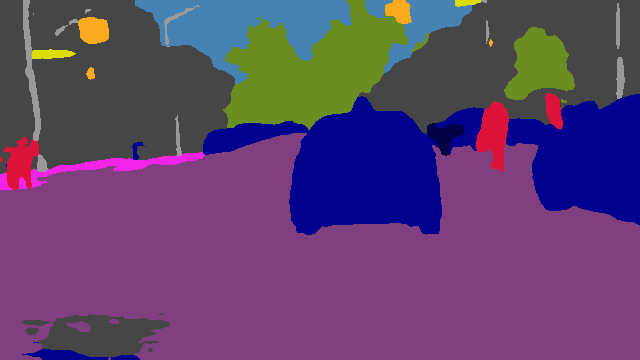}}
    \subfigure[Rainy Seg 2]{\includegraphics[width=0.24\columnwidth]{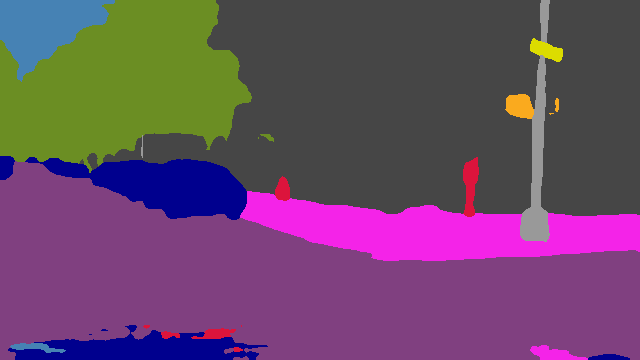}}
    \subfigure[Rainy Seg 3]{\includegraphics[width=0.24\columnwidth]{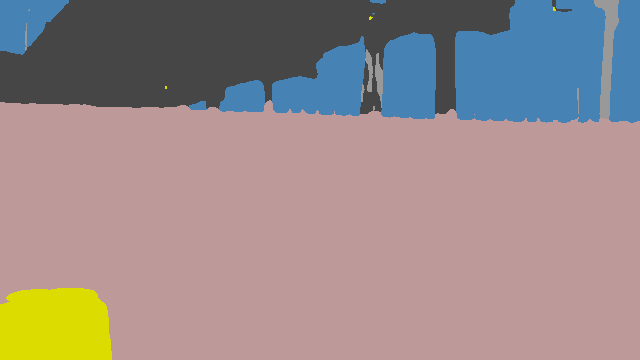}}
    \caption{\textbf{We prefer mPA and mIoU for segmentation maps over PSNR and SSIM for images in the context of regularizing contrastive learning.} For \textit{Clear image} paired with \textit{Rainy Image 1}, image scores are: [PSNR/SSIM] = [27.945/0.154] and segmentation scores are [mPA/mIoU] = [0.592/0.140]. For \textit{Clear image} paired with \textit{Rainy Image 2}, image scores are: [PSNR/SSIM] = [27.889/0.111] and for segmentation [mPA/mIoU] = [0.466/0.029]. Notably, mPA and mIoU capture semantic similarities in segmentation more effectively, with higher scores for more aligned pairs, whereas PSNR and SSIM show limited variance across scenarios. }

    
    \label{fig:SeNCE}
\end{figure}

    \subsection{Revisiting NCEs}

     \textbf{PatchNCE}
    Patch Noise Contrastive Estimation (PatchNCE) \cite{park2020contrastive} aims to maximize the mutual information between the corresponding input and output patches as below:
    
    \begin{equation}
    \mathcal{L}_{\text{PatchNCE}}(X, Z)=-\sum_{i=1}^N \log \frac{e^{\frac{x_i \cdot z_i}{\tau}}}{e^{\frac{x_i \cdot z_i}{\tau}}+\sum_{\substack{j=1 \\ j \neq i}}^N e^{\frac{x_i \cdot z_j }{\tau} }}
    \end{equation}
    
    Where $N$ is the number of patches, $\left[x_1, x_2, \cdots, x_N\right]$ and $\left[z_1, z_2, \cdots, z_N\right]$ are encoded patch features, and $\tau$ is the temperature hyperparameter.



     \textbf{MoNCE} The problem with PatchNCE \cite{park2020contrastive} is that it indiscriminately pushes all negative patches from the anchor, leading to sub-optimal performance for tasks with mixed easy and hard negatives patches. Modulated Noise Contrastive Estimation (MoNCE) \cite{zhan2022modulated} address this issue by reweighting the pushing force for the negative patches based on their similarity with the anchor as below.

    
    \begin{equation}
    \begin{aligned}
        &\mathcal{L}_{\text{MoNCE}}(X, Z)= \\ 
        &-\sum_{i=1}^N \log \frac{e^{\frac{x_i \cdot z_i}{\tau}}}{e^{\frac{x_i \cdot z_i}{\tau}}+Q(N-1) \sum_{\substack{j=1 \\ j \neq i}}^N w_{i j} \cdot e^{\frac{x_i \cdot z_j}{\tau}}}
    \end{aligned}
    \end{equation}
    
    Where $Q$ is a hyperparameter, and $w_{i j} (j \neq i)$ represents a weighting strategy. MoNCE proposes a hard weighting strategy $w_{i j}^{+}$ and an easy weighting strategy $w_{i j}^{-}$ to address unpaired and paired image-to-image translation, respectively. $w_{i j}^{+}$ and $w_{i j}^{-}$ are written as below.

    \begin{equation}
        w_{ij}^{+} = \text{softmax}\left(\frac{x_i \cdot z_j}{\beta}\right)_j 
        w_{ij}^{-} = \text{softmax}\left(\frac{1 - x_i \cdot z_j}{\beta}\right)_j
    \end{equation}

\begin{figure}[!htb]
    \centering
    \includegraphics[width=\columnwidth]{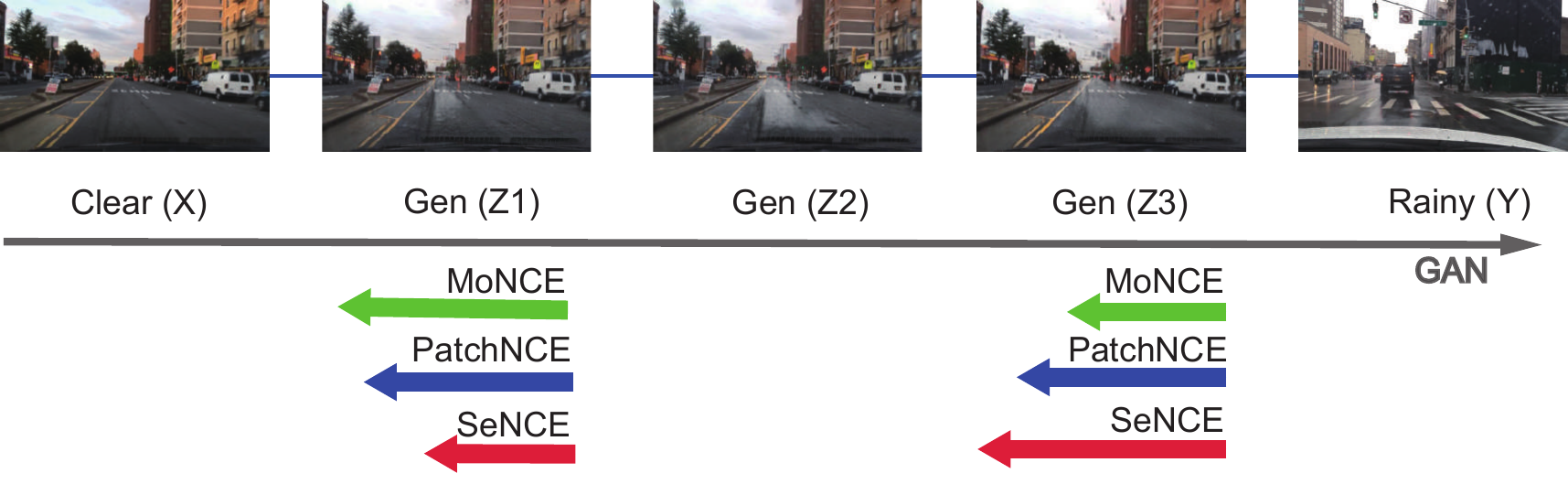}
    \caption{\textbf{SeNCE outperforms others in optimizing rain amount to produce realistic rainy images}. The length of the arrow here represents the magnitude of the NCE losses. }
    \label{fig:SeNCE_explain}
\end{figure}

    \subsection{Semantic Noise Contrastive Estimation (SeNCE)}

    While MoNCE \cite{zhan2022modulated} has improved upon PatchNCE \cite{park2020contrastive} on some benchmark image-to-image translation datasets, two issues impair its performance.

    First, MoNCE employs distinct weighting strategies for paired and unpaired settings without strong justification. In fact, in unpaired scenarios, images may appear paired, such as the same house from different angles or identical parking lots with different cars. In such cases, $w_{i j}$ should lean towards $w_{i j}^{-}$ rather than $w_{i j}^{+}$. To sum, a smooth transition between these weighting strategies would be more ideal.
    

    Second, MoNCE uses only image-level information from random patches for reweighting. Yet, in rain generation, many target domain pixels are compromised by droplets, streaks, wetness and mist. Such pixels can't provide precise guidance for contrastive learning \cite{chen2013generalized,chen2021benchmarks}. Hence, it's crucial to move beyond image-level details and seek a deeper understanding with minimal rain interference.
    

      It is evident from Fig. \ref{fig:SeNCE} that semantic-level metrics like mPA and mIoU more accurately capture the similarity between unpaired clear and rainy images than image-level metrics like PSNR and SSIM. This is because semantic-level metrics doesn't depend on perfectly aligned, corruption-free pixels. Even with unaligned images or pixels marred by rain, we can discern their differences using a comprehensive understanding of the segmentation maps. While we occasionally encounter pairs with very low mPA (e.g., 0.162 for clear and rainy img 3), such cases are rare and have minimal impact on training. Thus, we still include them in our training dataset. Below is our formulation for Semantic Noise Contrastive Estimation (SeNCE).

    \begin{equation}
    \label{eq:SeNCE}
    \begin{aligned}
        &\mathcal{L}_{\text{SeNCE}}(X, Y, Z)= \\ 
        &-\sum_{i=1}^N \log \frac{e^{\frac{x_i \cdot z_i}{\tau}}}{e^{\frac{x_i \cdot z_i}{\tau}}+Q(N-1) \sum_{\substack{j=1 \\ j \neq i}}^N w_{i j} \cdot e^{\frac{x_i \cdot z_j}{\tau}}}
    \end{aligned}
    \end{equation}
    
    Our weight $w_{i,j}$ in the above equation is:

    \begin{equation}
        \label{eq:wij}
        w_{ij} = \text{softmax}\left( \frac{F(i,j)}{\beta} \right)_j
    \end{equation}

    Where $F(i,j)$ can be expressed as:
        
    \begin{equation}
    \label{eq:Fij}
    \begin{aligned}
        &F(i, j) = \\ 
        &(1-m P A(X, Y))  (x_i \cdot z_j) + (m P A(X, Y))(1-x_i \cdot z_j)
    \end{aligned}
    \end{equation}

    
    $F(i,j)$ represents a semantic-based contrastive learning force derived from mPA. It adjusts between easy weights $w_{i j}^{-}$ and hard weights $w_{i j}^{+}$ of MoNCE. High mPA, indicating semantically similar clear and rainy images, results in a shift towards $w_{i j}^{-}$. In contrast, low mPA, indicating dissimilar images, leans towards hard weights $w_{i j}^{+}$.

    \subsection{Analysis of NCEs}

    Using the notation $Sim(X, Z) = x_i \cdot z_i$, we analyze the three NCEs with the help of Fig. \ref{fig:SeNCE_explain}:

    \begin{itemize}
        \item For insufficient rain ($Z_1$), $Sim(X,Z)$ does not align with $mPA(X,Y)$, and the weight $w_{ij}$ is large based on Eq. \ref{eq:wij} and Eq. \ref{eq:Fij} . According to Eq. \ref{eq:SeNCE}, this results in a low absolute value for $\mathcal{L}_{\text{SeNCE}}(X, Y, Z)$, allowing the GAN loss to dominate and drive the image towards the desired state ($Z_2$).
        \item For excessive rain ($Z_3$), $Sim(X,Z)$ closely matches $mPA(X,Y)$, and the weight $w_{ij}$ is small. This increases the magnitude of $\mathcal{L}_{\text{SeNCE}}(X, Y, Z)$, enabling SeNCE to overpower the GAN loss to guide the image back towards $Z_2$.
    \end{itemize}

    In essence, SeNCE adjusts the NCE loss based on the comparability of $mPA(X,Y)$ and $Sim(X,Z)$, refining the generated rain. Unlike PatchNCE, which lacks weight adjustment, and MoNCE, which doesn't account for semantic similarity, SeNCE ensures more realistic results.

    \subsection{Final Objective}
    The training objective of the proposed method is:
    
    \begin{equation}
    \begin{aligned}
    &\mathcal{L}_{(X, Y)} = \\ 
    &\lambda_1 \mathcal{L}_{\mathrm{GAN}}(X, Y) 
    + \lambda_2 \mathcal{L}_{\text {SeNCE}}(X, Y, Z) 
    + \lambda_3 \mathcal{L}_{\text {TPS}}(X, Y, Z)        
    \end{aligned}
    \end{equation}

    Similar to CUT \cite{park2020contrastive}, we set both $\lambda_1$ and $\lambda_2$ as 1. Since TPS is an auxiliary loss \cite{zheng2021deblur}, we set $\lambda_3$ as 0.1.

    \begin{table}[t]
    \centering
    \scriptsize
     \setlength\tabcolsep{3.0pt}
    \begin{tabular}{c|c|c|c|c|c}
    \hline
    \textbf{Variants}  & \textbf{TrainA}         & \textbf{TestA}         & \textbf{TrainB} & \textbf{TestB} & \textbf{Height x Width}                                               \\ \hline
    BDD100K (clear2rainy)  & \multirow{3}{*}{27,988} & \multirow{3}{*}{4,025} & 12,798          & 3,301          & \multirow{3}{*}{720 x 1,280}                                          \\ \cline{1-1} \cline{4-5}
    BDD100K (clear2snowy)  &                         &                        & 4,025           & 422            &                                                                       \\ \cline{1-1} \cline{4-5}
    BDD100K (day2night)    &                         &                        & 22,884           & 3,274          &                                                                       \\ \hline
    INIT (clear2rainy) & 18,112                  & 3,197                  & 3,330           & 588            & \begin{tabular}[c]{@{}c@{}}1,208 x 1,920\\ 3,000 x 4,000\end{tabular} \\ \hline

    Boreas (clear2snowy) & 8,356    & 2,089     & 3,649     & 913     & 720 x 860  \\ \hline
    
    \end{tabular}
    \caption{\textbf{Training and testing images for the task A2B.}}
    \label{tab:datasets}
    \end{table}

\section{Experiments}

\begin{table}[t]
\scriptsize
\centering
\setlength\tabcolsep{1.0pt}
\begin{tabular}{c|cc|cccc|cccc}
\hline
\multirow{3}{*}{\textbf{Model}} & \multicolumn{2}{c|}{\textbf{Constraints}}                                          & \multicolumn{4}{c|}{\textbf{NCEs}}                                                                                                                                                                                                                                                                         & \multicolumn{4}{c}{\textbf{Scores}}                                                                                                                                                               \\ \cline{2-11} 
                                   & \multicolumn{1}{c|}{\multirow{2}{*}{\textbf{PTL}}} & \multirow{2}{*}{\textbf{TPS}} & \multicolumn{1}{c|}{\multirow{2}{*}{\textbf{\begin{tabular}[c]{@{}c@{}}Patch- \\ NCE\end{tabular}}}} & \multicolumn{1}{c|}{\multirow{2}{*}{\textbf{\begin{tabular}[c]{@{}c@{}}Mo- \\ NCE\end{tabular}}}} & \multicolumn{1}{c|}{\multirow{2}{*}{\textbf{\begin{tabular}[c]{@{}c@{}}SeNCE \\ (mIoU)\end{tabular}}}} & \multirow{2}{*}{\textbf{\begin{tabular}[c]{@{}c@{}}SeNCE \\ (mPA)\end{tabular}}} & \multicolumn{1}{c|}{\multirow{2}{*}{\textbf{Content$\uparrow$}}} & \multicolumn{1}{c|}{\multirow{2}{*}{\textbf{Style$\uparrow$}}} & \multicolumn{1}{c|}{\multirow{2}{*}{\textbf{KID$\downarrow$}}} & \multirow{2}{*}{\textbf{FID$\downarrow$}} \\
                                   & \multicolumn{1}{c|}{}                              &                               & \multicolumn{1}{c|}{}                                   & \multicolumn{1}{c|}{}                                & \multicolumn{1}{c|}{}                                                                                  &                                                                                  & \multicolumn{1}{c|}{}                                  & \multicolumn{1}{c|}{}                                & \multicolumn{1}{c|}{}                              &                               \\ \hline
M1                                 & \multicolumn{1}{c|}{}                              &                               & \multicolumn{1}{c|}{$\checkmark$}                                & \multicolumn{1}{c|}{}                                & \multicolumn{1}{c|}{}                                                                                  &                                                                                  & \multicolumn{1}{c|}{3.32}                              & \multicolumn{1}{c|}{3.38}                            & \multicolumn{1}{c|}{85.29}                         & 21.90                         \\ \hline
M2                                 & \multicolumn{1}{c|}{$\checkmark$}                           &                               & \multicolumn{1}{c|}{$\checkmark$}                                & \multicolumn{1}{c|}{}                                & \multicolumn{1}{c|}{}                                                                                  &                                                                                  & \multicolumn{1}{c|}{2.93}                              & \multicolumn{1}{c|}{2.83}                            & \multicolumn{1}{c|}{88.15}                         & 22.12                         \\ 
M3                                 & \multicolumn{1}{c|}{}                              & $\checkmark$                           & \multicolumn{1}{c|}{$\checkmark$}                                & \multicolumn{1}{c|}{}                                & \multicolumn{1}{c|}{}                                                                                  &                                                                                  & \multicolumn{1}{c|}{3.51}                              & \multicolumn{1}{c|}{{\color{blue}{3.58}}}                      & \multicolumn{1}{c|}{\textbf{70.93}}                & 20.90                         \\ \hline
M4                                 & \multicolumn{1}{c|}{\textbf{}}                     & \textbf{}                     & \multicolumn{1}{c|}{\textbf{}}                          & \multicolumn{1}{c|}{$\checkmark$}                             & \multicolumn{1}{c|}{}                                                                                  &                                                                                  & \multicolumn{1}{c|}{3.10}                              & \multicolumn{1}{c|}{3.40}                            & \multicolumn{1}{c|}{75.66}                         & \textbf{18.60}                \\ 
M5                                 & \multicolumn{1}{c|}{}                              &                               & \multicolumn{1}{c|}{}                                   & \multicolumn{1}{c|}{}                                & \multicolumn{1}{c|}{}                                                                                  & $\checkmark$                                                                              & \multicolumn{1}{c|}{{\color{blue}{3.52}}}                        & \multicolumn{1}{c|}{3.56}                            & \multicolumn{1}{c|}{74.20}                         & 20.64                         \\  \hline
M6                                 & \multicolumn{1}{c|}{}                              & $\checkmark$                           & \multicolumn{1}{c|}{}                                   & \multicolumn{1}{c|}{}                                & \multicolumn{1}{c|}{$\checkmark$}                                                                               &                                                                                  & \multicolumn{1}{c|}{3.27}                              & \multicolumn{1}{c|}{3.33}                            & \multicolumn{1}{c|}{80.37}                         & 21.12                         \\ 
M7                                 & \multicolumn{1}{c|}{}                              & $\checkmark$                           & \multicolumn{1}{c|}{}                                   & \multicolumn{1}{c|}{}                                & \multicolumn{1}{c|}{}                                                                                  & $\checkmark$                                                                              & \multicolumn{1}{c|}{\textbf{3.58}}                     & \multicolumn{1}{c|}{\textbf{3.70}}                   & \multicolumn{1}{c|}{{\color{blue}{}72.19}}                   & {\color{blue}{19.34}}                   \\ \hline
\end{tabular}
\caption{\textbf{Quantitative ablation for rain generation on BDD100K dataset.} The best scores are in \textbf{bold}, and the second best scores are in \color{blue}{blue}.  }
\label{tab:ablation}
\end{table}

        \begin{figure}
            \centering

            \subfigure[Clear]{\includegraphics[width=0.24\columnwidth]{LaTeX/Vis_Compare/Rainmake_BDD/Clear/test/b1c66a42-6f7d68ca.jpg}} 
            \subfigure[M1]{\includegraphics[width=0.24\columnwidth]{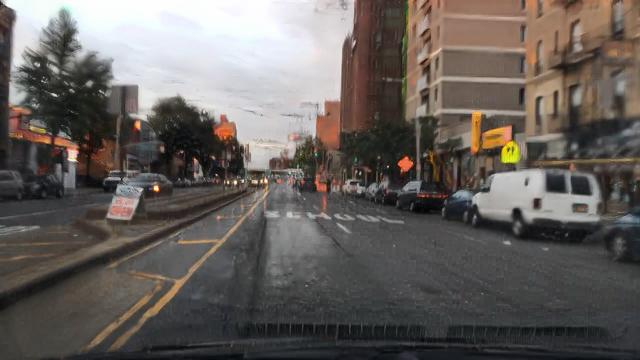}} 
            \subfigure[M2]{\includegraphics[width=0.24\columnwidth]{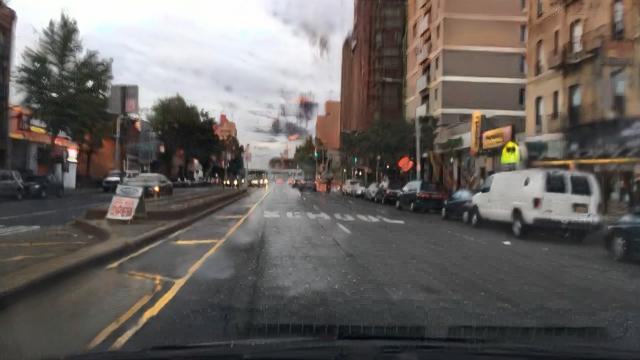}}
            \subfigure[M3]{\includegraphics[width=0.24\columnwidth]{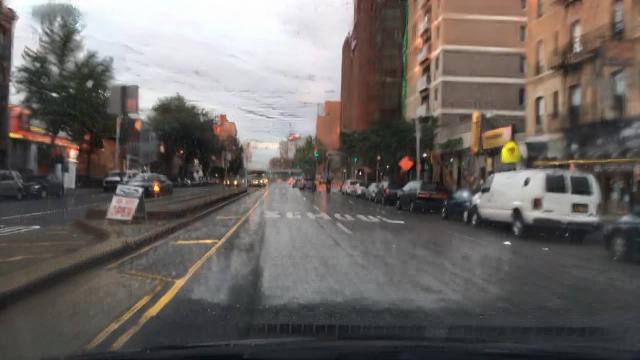}} \\

            \vspace{-0.2cm}

            \subfigure[M4]{\includegraphics[width=0.24\columnwidth]{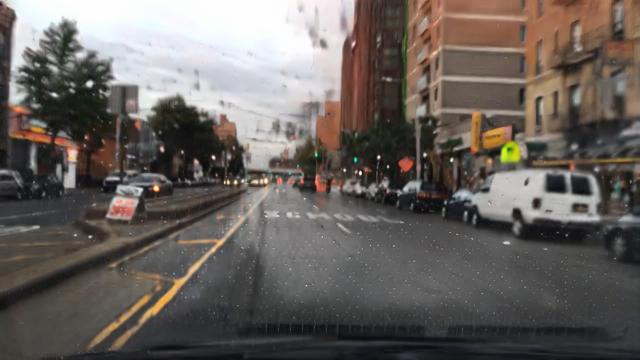}}
             \subfigure[M5]{\includegraphics[width=0.24\columnwidth]{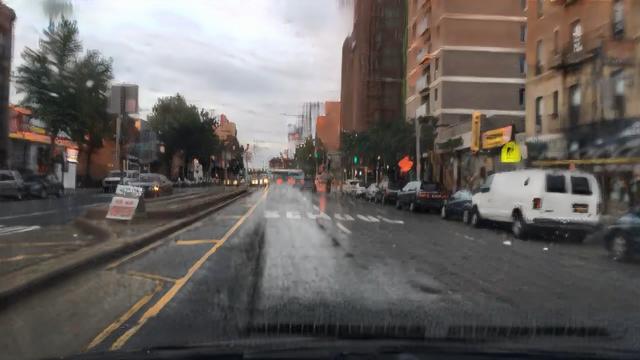}}
             \subfigure[M6]{\includegraphics[width=0.24\columnwidth]{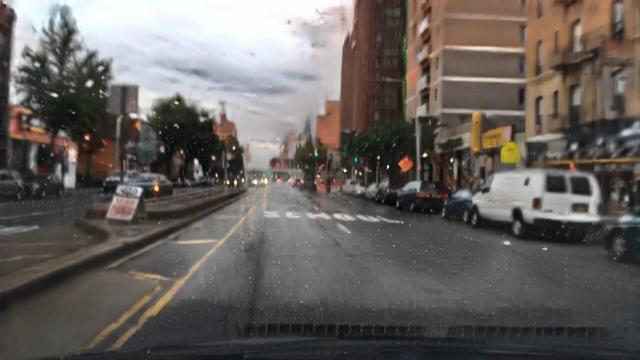}}
              \subfigure[M7]{\includegraphics[width=0.24\columnwidth]{LaTeX/Vis_Compare/Rainmake_BDD/CUT_TPS_SeNCE/test/b1c66a42-6f7d68ca.jpg}}

            \caption{\textbf{Qualitative ablation for rain generation on BDD100K dataset.} }
            \label{fig:vis_ablation}
        \end{figure}

\subsection{Implementation Details and Metrics}

\begin{table}[t]
 \setlength\tabcolsep{1.5pt}
\scriptsize
\centering
\begin{tabular}{c|cccc|cccc}
\hline
\multirow{2}{*}{\textbf{Methods}} & \multicolumn{4}{c|}{\textbf{BDD100K Dataset  (clear $\rightarrow$ rainy)}}                                                              & \multicolumn{4}{c}{\textbf{INIT Dataset (clear $\rightarrow$ rainy)}}                                                               \\ \cline{2-9} 
                                  & \multicolumn{1}{c|}{\textbf{Content$\uparrow$}} & \multicolumn{1}{c|}{\textbf{Style$\uparrow$}} & \multicolumn{1}{c|}{\textbf{KID$\downarrow$}}   & \textbf{FID$\downarrow$}    & \multicolumn{1}{c|}{\textbf{Content$\uparrow$}} & \multicolumn{1}{c|}{\textbf{Style$\uparrow$}} & \multicolumn{1}{c|}{\textbf{MMD$\downarrow$}}    & \textbf{ED$\downarrow$}     \\ \hline
UNIT                              & \multicolumn{1}{c|}{3.24}             & \multicolumn{1}{c|}{3.48}           & \multicolumn{1}{c|}{88.85}          & \textbf{18.099} & \multicolumn{1}{c|}{2.58}             & \multicolumn{1}{c|}{2.66}           & \multicolumn{1}{c|}{34.231}          & 35.702          \\ 
MUNIT                             & \multicolumn{1}{c|}{2.44}             & \multicolumn{1}{c|}{2.80}           & \multicolumn{1}{c|}{189.12}         & 26.538          & \multicolumn{1}{c|}{2.80}             & \multicolumn{1}{c|}{2.72}           & \multicolumn{1}{c|}{34.425}          & 36.458          \\ 
CUT                               & \multicolumn{1}{c|}{3.32}             & \multicolumn{1}{c|}{3.38}           & \multicolumn{1}{c|}{85.29}          & 21.901          & \multicolumn{1}{c|}{3.16}             & \multicolumn{1}{c|}{2.90}           & \multicolumn{1}{c|}{33.704}          & 34.777          \\ 
QS-Attn                           & \multicolumn{1}{c|}{3.34}             & \multicolumn{1}{c|}{3.58}           & \multicolumn{1}{c|}{85.59}          & 21.614          & \multicolumn{1}{c|}{2.46}             & \multicolumn{1}{c|}{2.66}           & \multicolumn{1}{c|}{33.836}          & 34.853          \\ 
MoNCE                             & \multicolumn{1}{c|}{3.10}             & \multicolumn{1}{c|}{3.30}           & \multicolumn{1}{c|}{75.66}          & 18.595          & \multicolumn{1}{c|}{2.18}             & \multicolumn{1}{c|}{2.24}           & \multicolumn{1}{c|}{33.579}          & 34.814          \\ 
Ours                              & \multicolumn{1}{c|}{\textbf{3.58}}    & \multicolumn{1}{c|}{\textbf{3.70}}  & \multicolumn{1}{c|}{\textbf{72.19}} & 19.341          & \multicolumn{1}{c|}{\textbf{3.42}}    & \multicolumn{1}{c|}{\textbf{3.04}}  & \multicolumn{1}{c|}{\textbf{33.535}} & \textbf{34.774} \\ \hline
\end{tabular}
\caption{\textbf{Quantitative comparison for image rain generation on the BDD100K and INIT datasets.} }
\label{tab:quant_bdd_init}
\end{table}

\begin{table}[t]
\centering
\scriptsize
\setlength\tabcolsep{0.1pt}
\begin{tabular}{c|cc|cc|cc|cc|cc}
\hline
\multirow{2}{*}{\begin{tabular}[c]{@{}c@{}}\textbf{Methods} \\  \end{tabular}} & \multicolumn{2}{c|} {\textbf{EffDerain}}              & \multicolumn{2}{c|}{\textbf{VRGNet}}                 & \multicolumn{2}{c|}{\textbf{PreNet}}                 & \multicolumn{4}{c}{\textbf{SAPNet}}                \\ \cline{2-11} 
                                                                                                    & \multicolumn{1}{c|}{\textbf{DBCNN}}          & \textbf{MUSIQ}          & \multicolumn{1}{c|}{\textbf{DBCNN}}          & \textbf{MUSIQ}          & \multicolumn{1}{c|}{\textbf{DBCNN}}          & \textbf{MUSIQ}          & \multicolumn{1}{c|}{\textbf{DBCNN}}          & \textbf{MUSIQ}          & \multicolumn{1}{c|}{\textbf{Qual}}          & \textbf{Perf}          \\ \hline
Rain100H                                                                                            & \multicolumn{1}{c|}{39.33}          & 49.51          & \multicolumn{1}{c|}{36.10}          & 49.65          & \multicolumn{1}{c|}{47.34}          & 53.72          & \multicolumn{1}{c|}{46.59}          & 53.27          & \multicolumn{1}{c|}{3.20}           & 2.08           \\ 
UNIT                                                                                                & \multicolumn{1}{c|}{36.56}          & 48.73          & \multicolumn{1}{c|}{43.07}          & 51.73          & \multicolumn{1}{c|}{44.16}          & 55.05          & \multicolumn{1}{c|}{45.91}          & 56.04          & \multicolumn{1}{c|}{3.37}           & 3.49           \\ 
MUNIT                                                                                               & \multicolumn{1}{c|}{38.54}          & 52.55          & \multicolumn{1}{c|}{37.22}          & 50.21          & \multicolumn{1}{c|}{37.44}          & 54.28          & \multicolumn{1}{c|}{41.92}          & 55.07          & \multicolumn{1}{c|}{3.04}           & 2.20           \\ 
CUT                                                                                                 & \multicolumn{1}{c|}{36.34}          & 48.82          & \multicolumn{1}{c|}{37.22}          & 50.21          & \multicolumn{1}{c|}{55.00}          & 60.67          & \multicolumn{1}{c|}{54.49}          & 59.97          & \multicolumn{1}{c|}{2.88}           & 3.92          \\ 
QS-Attn                                                                                             & \multicolumn{1}{c|}{37.02}          & 49.62          & \multicolumn{1}{c|}{36.93}          & 50.81          & \multicolumn{1}{c|}{34.61}          & 43.02          & \multicolumn{1}{c|}{57.31}          & 61.06          & \multicolumn{1}{c|}{2.94}           & 3.22            \\
MoNCE                                                                                               & \multicolumn{1}{c|}{37.30}          & 49.63          & \multicolumn{1}{c|}{37.44}          & 50.96          & \multicolumn{1}{c|}{59.49}          & 60.68          & \multicolumn{1}{c|}{37.08}          & 48.95          & \multicolumn{1}{c|}{2.78}           & 2.37           \\
Ours                                                                                                & \multicolumn{1}{c|}{\textbf{39.82}} & \textbf{54.51} & \multicolumn{1}{c|}{\textbf{47.87}} & \textbf{54.10} & \multicolumn{1}{c|}{\textbf{60.17}} & \textbf{61.79} & \multicolumn{1}{c|}{\textbf{57.83}} & \textbf{61.85} & \multicolumn{1}{c|}{\textbf{3.53}}           & \textbf{4.33}          \\ \hline
\end{tabular}
\caption{\textbf{Deraining comparisons with different deraining methods trained on images from different rain generation methods}. For all metrics in this table, higher is better. }
\label{tab:quant_derain}
\end{table}

    \begin{table}[t]
    \scriptsize
    \setlength\tabcolsep{1.0pt}
    \centering
    \begin{tabular}{c|c|c|c|c|c|c|c}
    \hline
    \textbf{Pretrained}   & \textbf{Finetuned} & \textbf{mAP}   & \textbf{mAP\_50} & \textbf{mAP\_75} & \textbf{mAP\_s} & \textbf{mAP\_m} & \textbf{mAP\_l} \\ \hline
    \multirow{8}{*}{COCO} & None               & 0.171          & 0.373            & 0.134            & 0.093           & 0.300           & 0.409           \\ 
                          & Clear              & 0.231          & 0.492            & 0.190            & 0.151           & 0.366           & 0.454           \\ 
                          & UNIT               & 0.215          & 0.462            & 0.178            & 0.113           & 0.404           & 0.546           \\ 
                          & MUNIT              & 0.178          & 0.394            & 0.147            & 0.108           & 0.291           & 0.438           \\ 
                          & CUT                & 0.243          & 0.512            & 0.203            & 0.149           & 0.410           & 0.548           \\ 
                          & QS-Attn            & 0.248          & 0.513            & 0.213            & 0.151           & 0.401           & 0.574           \\ 
                          & MoNCE              & 0.246          & 0.510            & 0.211            & 0.151           & 0.400           & 0.570           \\ 
                          & Ours               & \textbf{0.262} & \textbf{0.526}   & \textbf{0.237}   & \textbf{0.158}  & \textbf{0.419}  & \textbf{0.646}  \\ \hline
    \end{tabular}
    \caption{\textbf{Quantitative comparison for Yolov3 object detection on the BDD100K test rainy dataset.} The Yolov3 object detector are pretrained on COCO \cite{lin2014microsoft}, and finetuned on generated rainy images using different  methods. mAP is computed on the most challenging 100 rainy images. }    \label{tab:obj_det_rainmake}
    \end{table}

\textbf{Implementation Details:}
We trained our model with PyTorch \cite{paszke2019pytorch} on 8 RTX 3090 GPUs with an Intel Xeon Gold 6330 Processor CPU, using the Adam optimizer \cite{kingma2014adam}. The model was trained for 200 epochs, with a learning rate of $2e^{-4}$ for the first 100 epochs, decreased to $2e^{-5}$ for the remaining epochs. Training utilized randomly cropped patches of size 256 $\times$ 256. All rain generation baselines were trained under that setting for fair comparison.\footnote{The implementation details for benchmark datasets, as well as deraining and detection algorithms, can be found in the supplementary material.}



\textbf{Evaluation Metrics:}
We evaluated rain generation models using FID \cite{heusel2017gans} and KID \cite{binkowski2018demystifying} on the BDD100K and Boreas datasets. We employed MMD \cite{gretton2012kernel} and ED \cite{rizzo2016energy} for the INIT dataset, due to its small size \cite{salimans2016improved}. In all tables, KID values are scaled by $10^{-4}$ and MMD by $10^{-5}$. Additionally, a 50-participant user study rated generated image content and style on a 1-5 scale. For deraining, we train multiple deraining methods \cite{guo2021efficientderain, wang2021rain, ren2019progressive, zheng2022sapnet} on images generated from different rain generation models, evaluated them using MUSIQ \cite{ke2021musiq} and DBCNN \cite{hou2014blind}, and conducted another user study  for image quality of the derained images and the performance of the deraining methods. For detection, we used Yolov3 \cite{redmon2018yolov3}, evaluating performance with mAP across different IoU thresholds and object sizes.


        \begin{figure*}[t]
            \centering
            \includegraphics[width=2.4cm]{LaTeX/Vis_Compare/Rainmake_BDD/Clear/test/b1c66a42-6f7d68ca.jpg}
            \includegraphics[width=2.4cm]{LaTeX/Vis_Compare/Rainmake_BDD/CUT_TPS_SeNCE/test/b1c66a42-6f7d68ca.jpg}
            \includegraphics[width=2.4cm]{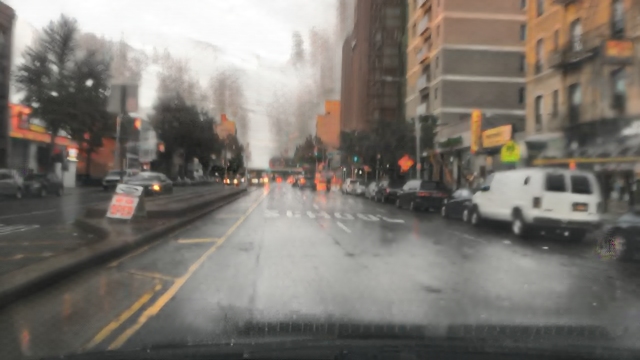}
             \includegraphics[width=2.4cm]{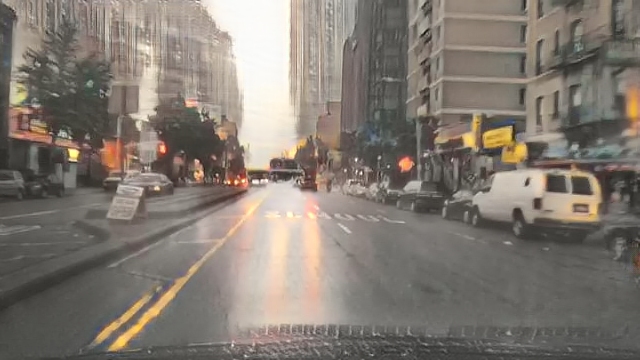}
              \includegraphics[width=2.4cm]{LaTeX/Vis_Compare/Rainmake_BDD/CUT/test/b1c66a42-6f7d68ca.jpg}
             \includegraphics[width=2.4cm]{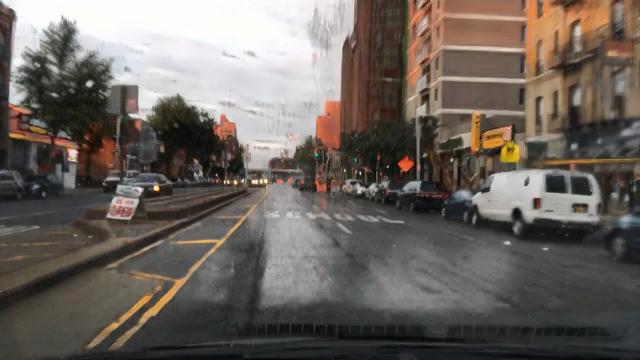}
            \includegraphics[width=2.4cm]{LaTeX/Vis_Compare/Rainmake_BDD/MoNCE/test/b1c66a42-6f7d68ca.jpg} \\

            \includegraphics[width=2.4cm]{LaTeX/Vis_Compare/Rainmake_BDD/Clear/test/b1d10d08-c35503b8.jpg}
            \includegraphics[width=2.4cm]{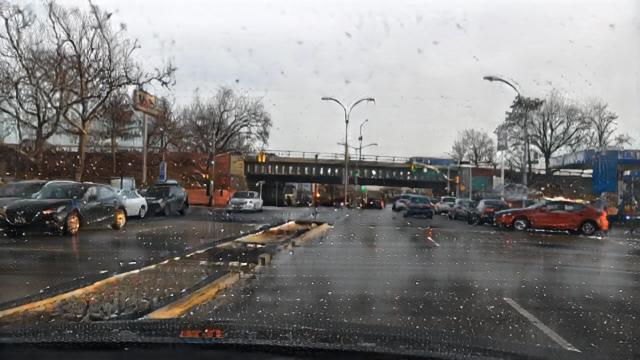}
            \includegraphics[width=2.4cm]{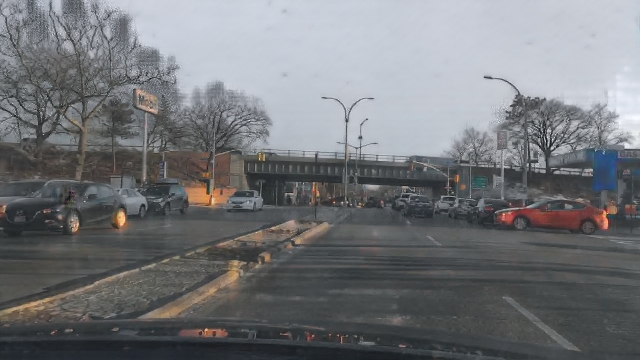}
             \includegraphics[width=2.4cm]{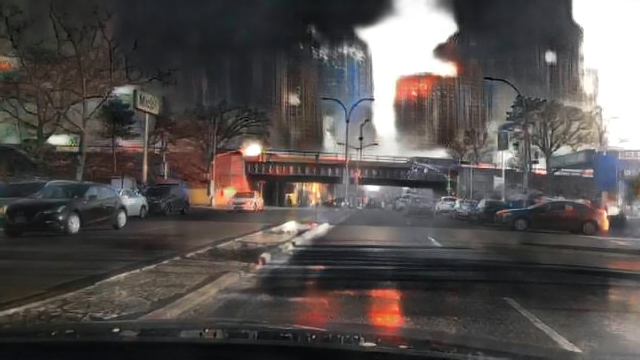}
              \includegraphics[width=2.4cm]{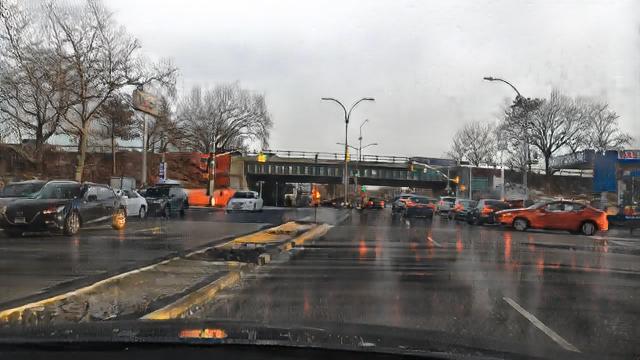}
             \includegraphics[width=2.4cm]{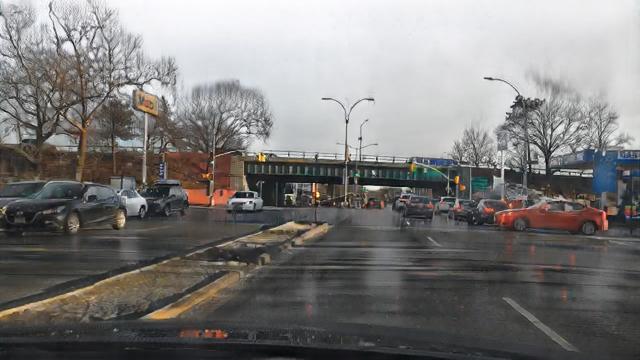}
            \includegraphics[width=2.4cm]{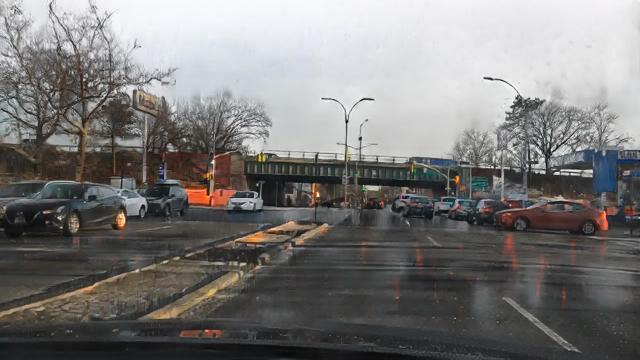} \\
            
            \includegraphics[width=2.4cm]{LaTeX/Vis_Compare/Rainmake_BDD/Clear/test/b1d10d08-5b108225.jpg}
            \includegraphics[width=2.4cm]{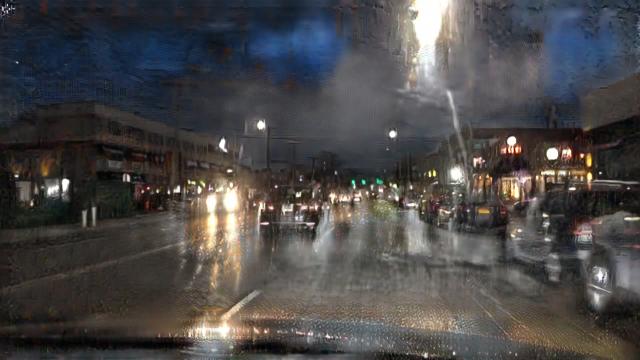}
            \includegraphics[width=2.4cm]{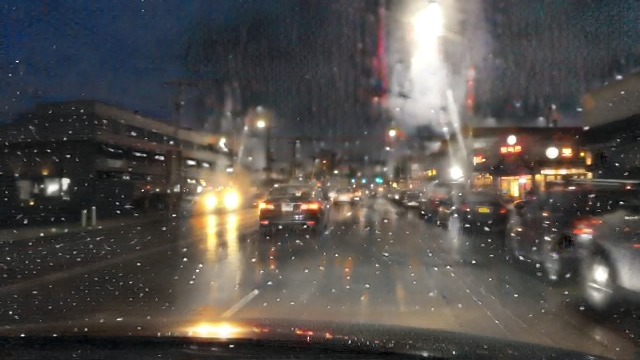}
             \includegraphics[width=2.4cm]{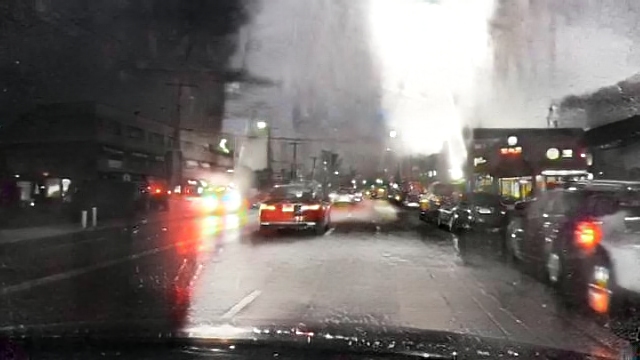}
              \includegraphics[width=2.4cm]{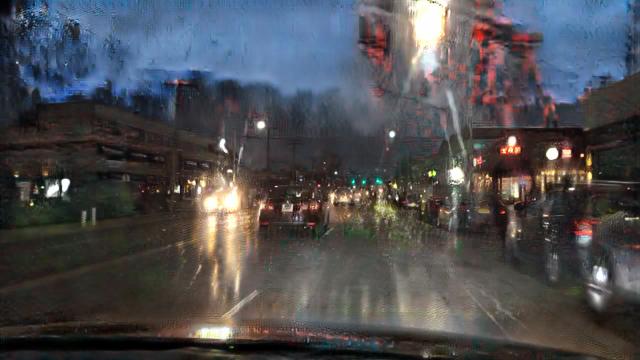}
             \includegraphics[width=2.4cm]{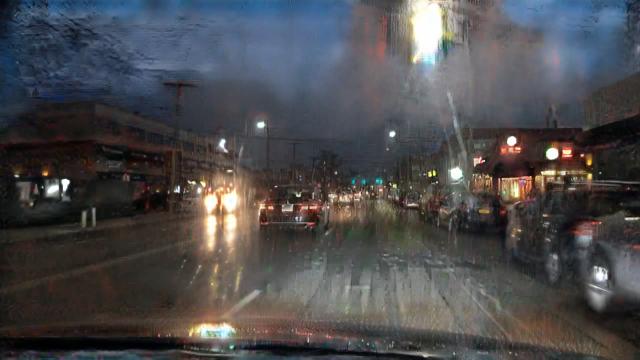}
            \includegraphics[width=2.4cm]{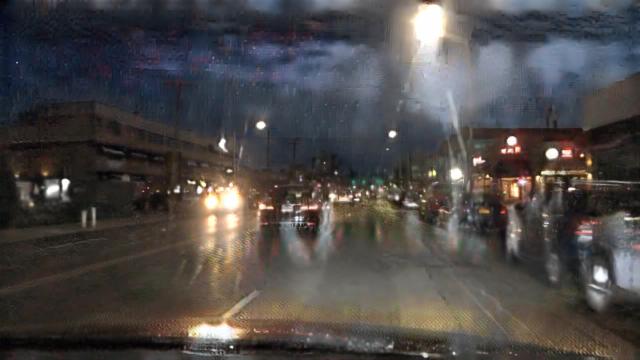} \\

            \vspace{-0.15cm}

            \subfigure[Clear]{\includegraphics[width=2.4cm]{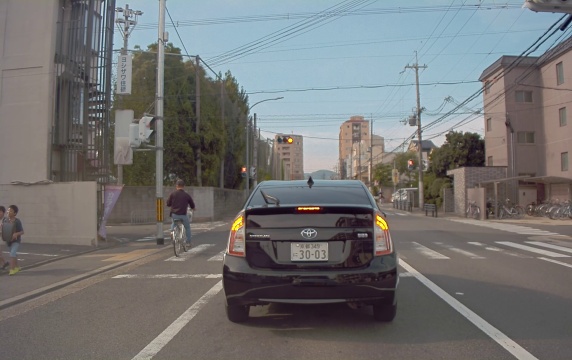}} 
             \subfigure[Ours]{\includegraphics[width=2.4cm]{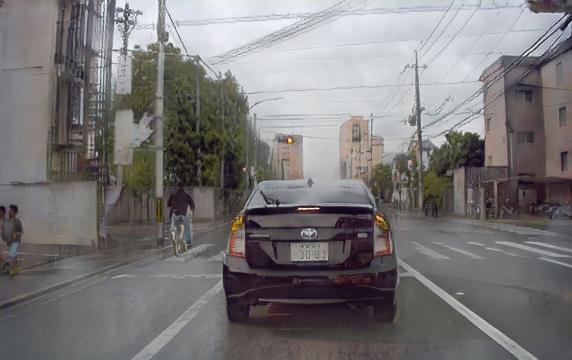}}
            \subfigure[UNIT \cite{liu2017unsupervised}]{\includegraphics[width=2.4cm]{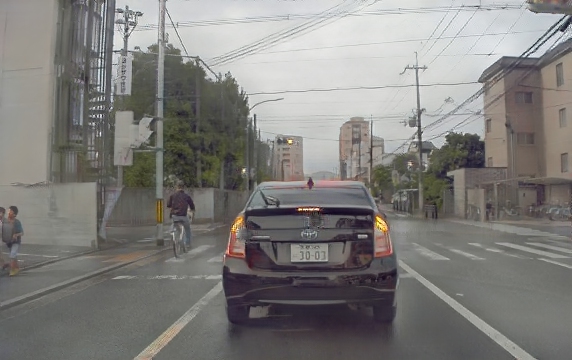}}
             \subfigure[MUNIT \cite{huang2018multimodal}]{\includegraphics[width=2.4cm]{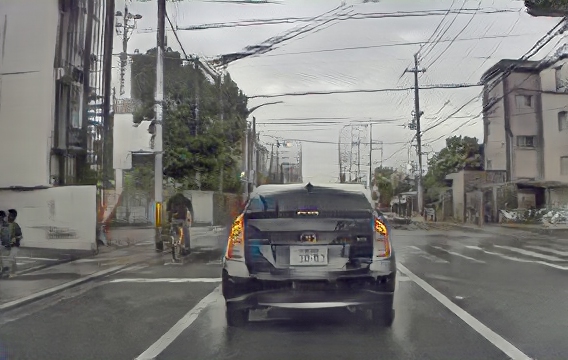}}
            \subfigure[CUT \cite{park2020contrastive}]{\includegraphics[width=2.4cm]{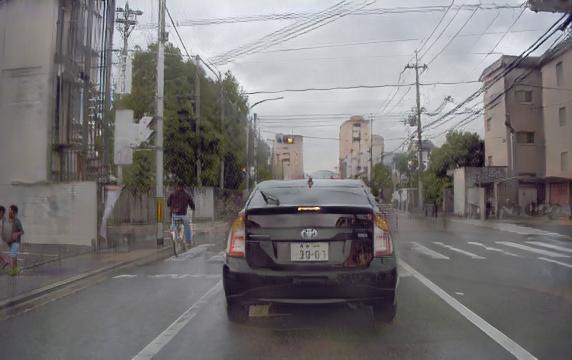}} 
            \subfigure[QS-Attn \cite{hu2022qs}]{\includegraphics[width=2.4cm]{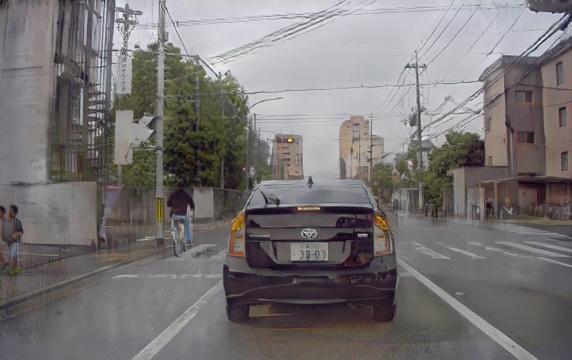}}
            \subfigure[MoNCE \cite{zhan2022modulated}]{\includegraphics[width=2.4cm]{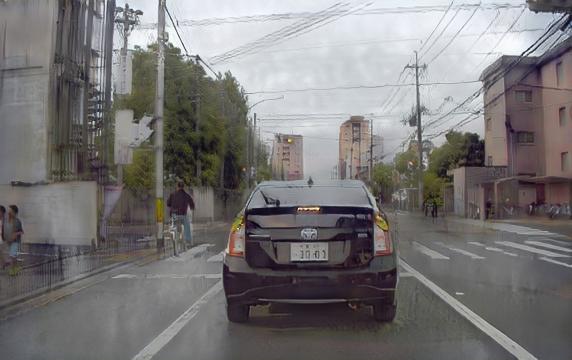}}
            \caption{\textbf{Qualitative comparison for clear2rainy (i.e. rain generation) on the BDD100K and INIT dataset.} }
            \label{fig:vis_raingen}
        \end{figure*}

        \begin{figure*}[t]

        
            \centering


            \includegraphics[width=2.4cm]{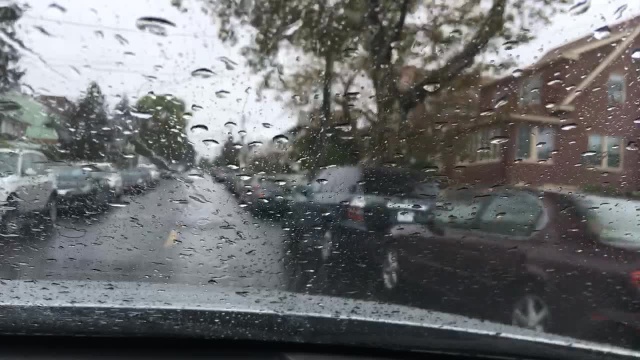}
            \includegraphics[width=2.4cm]{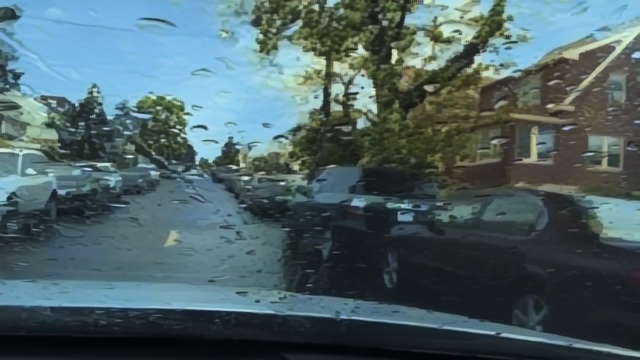}
            \includegraphics[width=2.4cm]{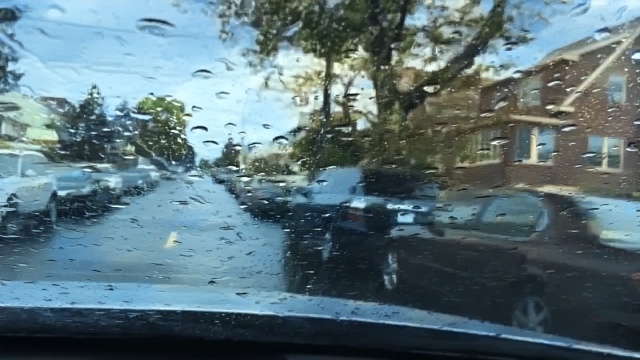}
             \includegraphics[width=2.4cm]{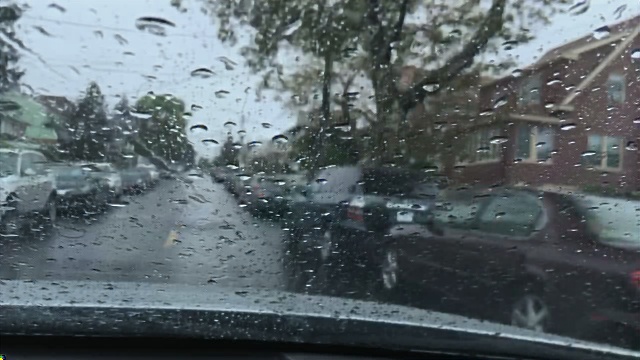}
              \includegraphics[width=2.4cm]{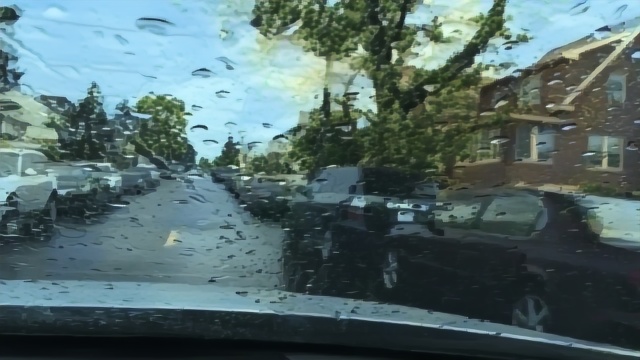}
             \includegraphics[width=2.4cm]{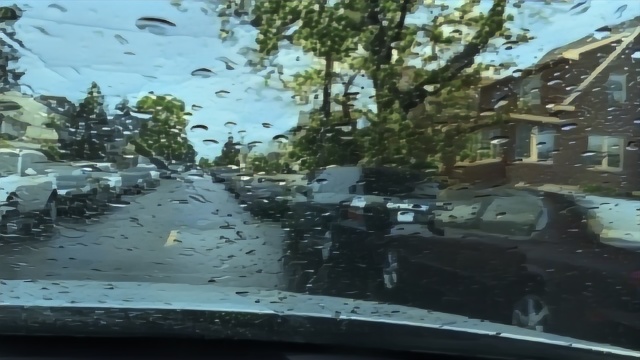}
            \includegraphics[width=2.4cm]{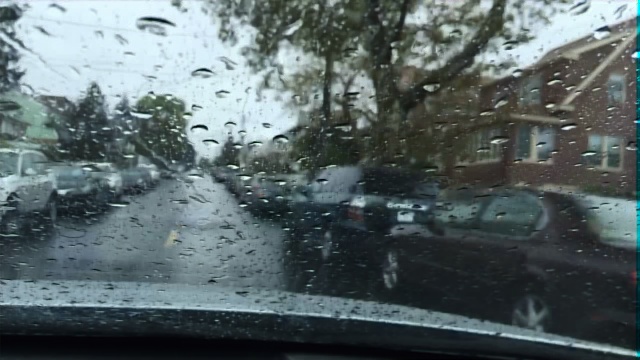} 
            \\

            \includegraphics[width=2.4cm]{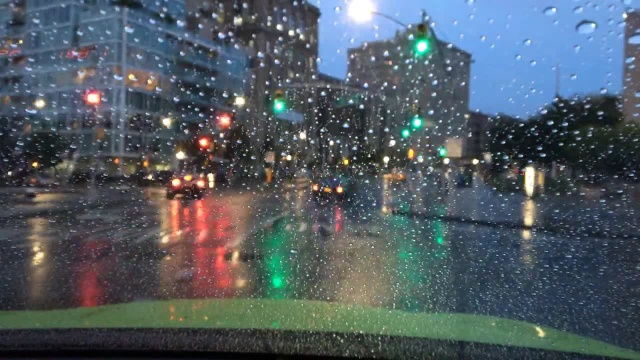}
            \includegraphics[width=2.4cm]{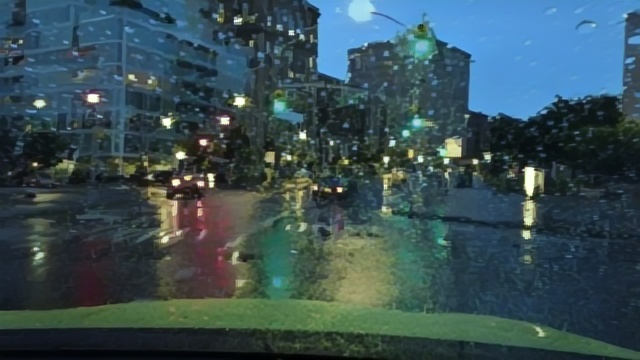}
            \includegraphics[width=2.4cm]{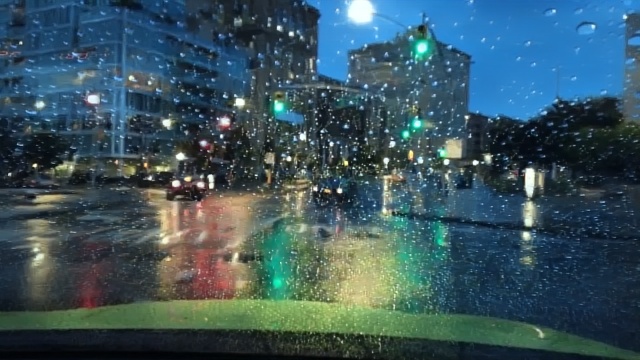}
             \includegraphics[width=2.4cm]{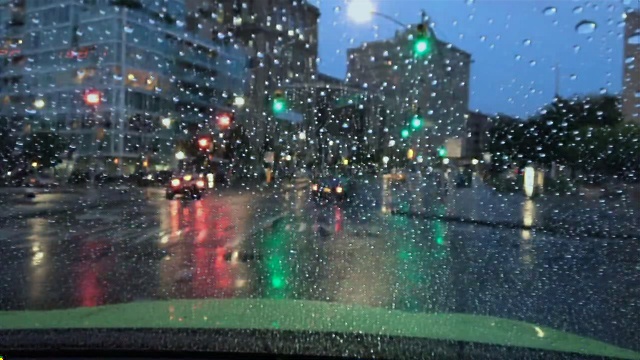}
              \includegraphics[width=2.4cm]{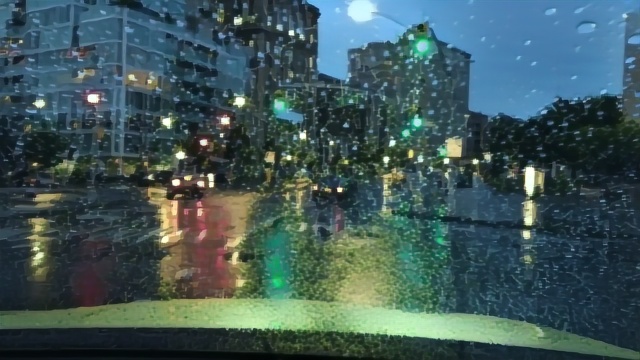}
             \includegraphics[width=2.4cm]{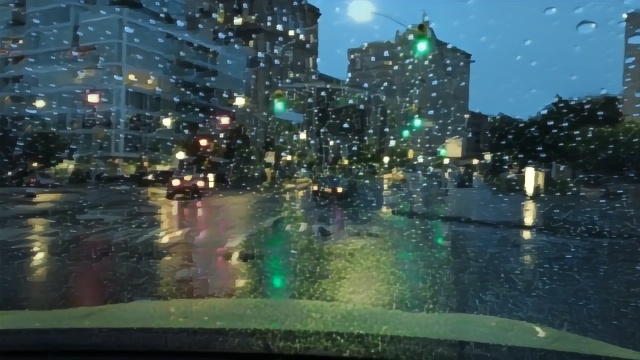}
            \includegraphics[width=2.4cm]{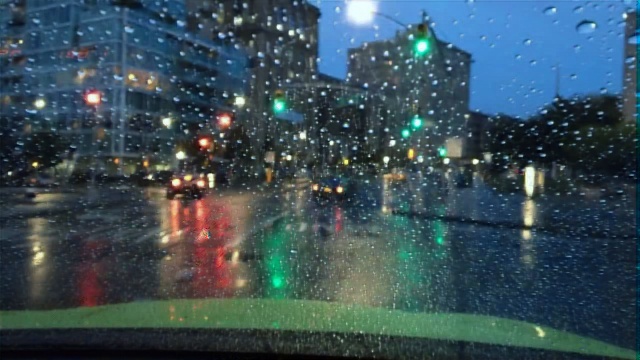} 
            \\

           \includegraphics[width=2.4cm]{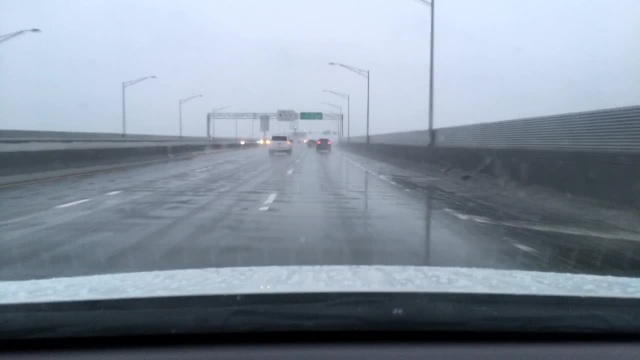}
            \includegraphics[width=2.4cm]{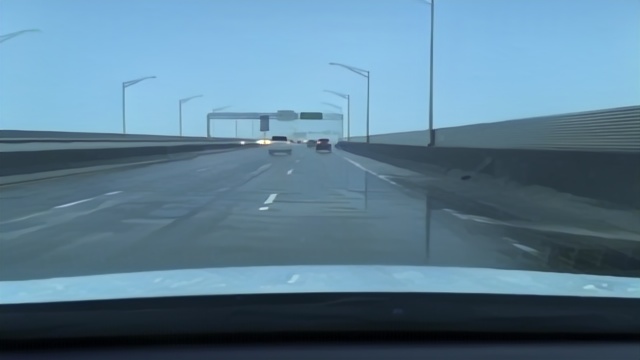}
            \includegraphics[width=2.4cm]{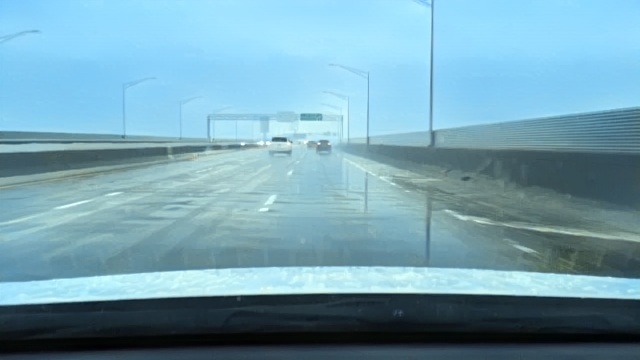}
             \includegraphics[width=2.4cm]{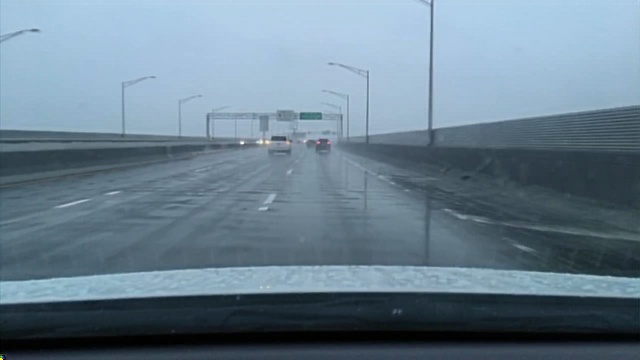}
              \includegraphics[width=2.4cm]{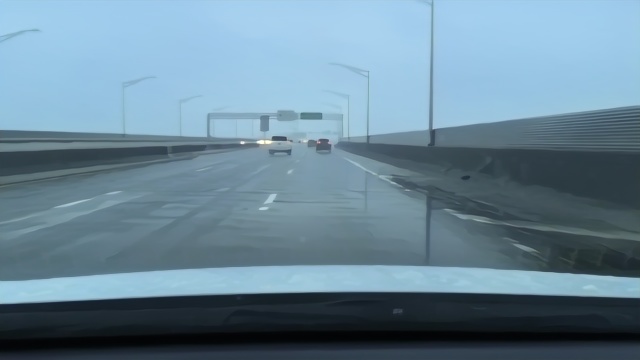}
             \includegraphics[width=2.4cm]{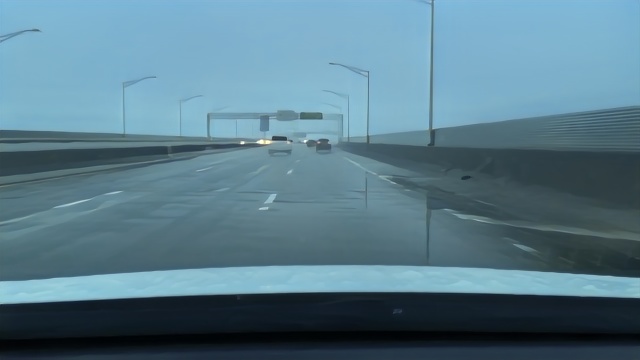}
            \includegraphics[width=2.4cm]{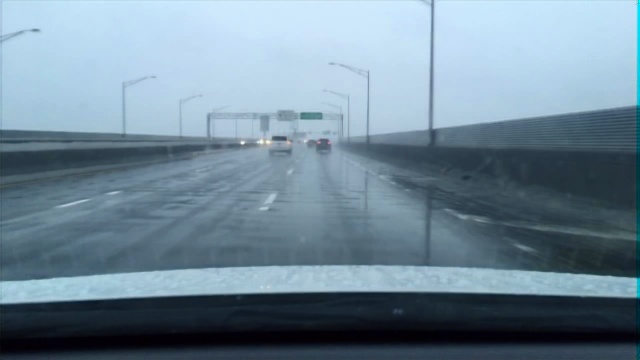} 
            \\

            \vspace{-0.15cm}


            \subfigure[Rainy]{\includegraphics[width=2.4cm]{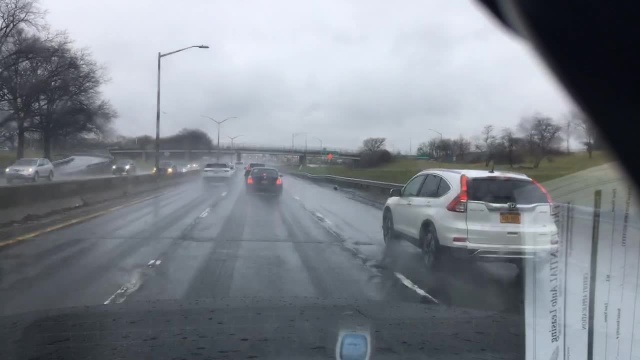}} 
            \subfigure[Ours]{\includegraphics[width=2.4cm]{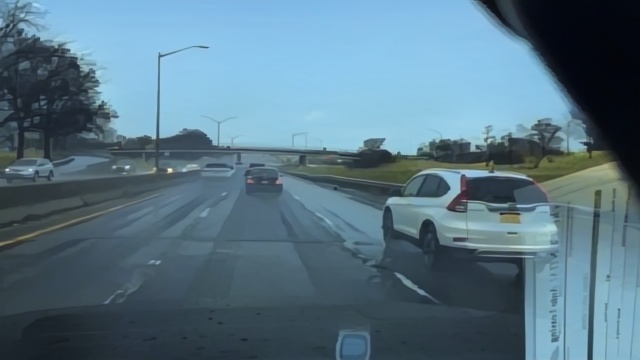}}
            \subfigure[UNIT \cite{liu2017unsupervised}]{\includegraphics[width=2.4cm]{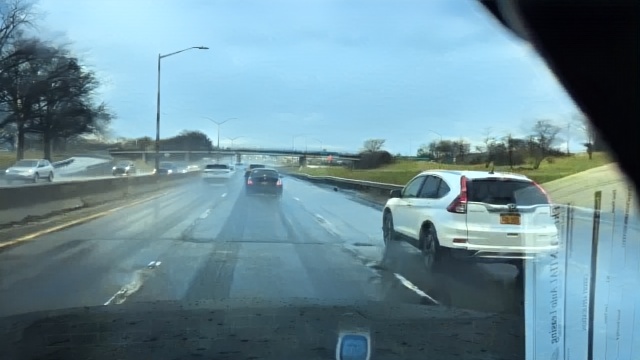}}
             \subfigure[MUNIT \cite{huang2018multimodal}]{\includegraphics[width=2.4cm]{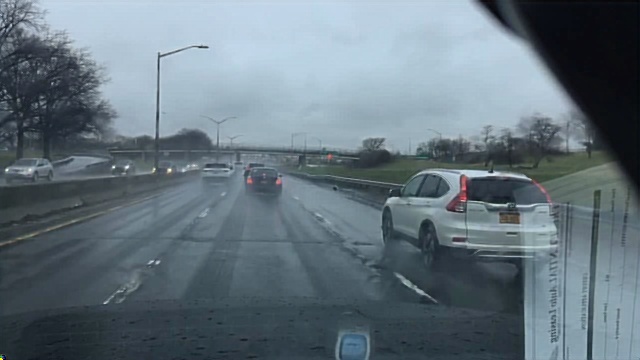}}
            \subfigure[CUT \cite{park2020contrastive}]{\includegraphics[width=2.4cm]{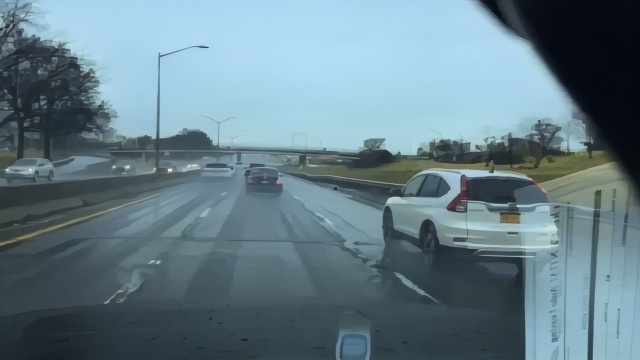}} 
            \subfigure[QS-Attn \cite{hu2022qs}]{\includegraphics[width=2.4cm]{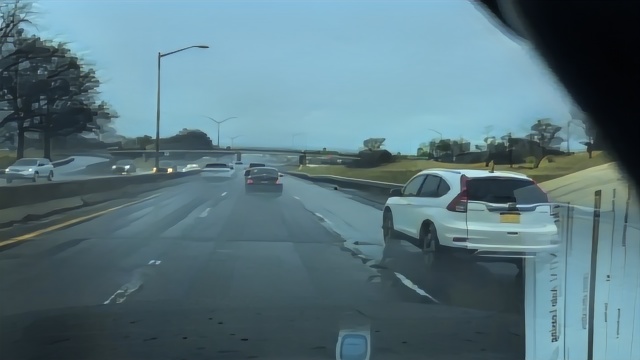}}
            \subfigure[MoNCE \cite{zhan2022modulated}]{\includegraphics[width=2.4cm]{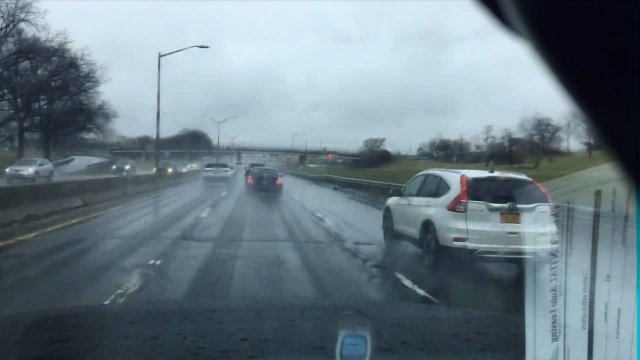}}
            \caption{\textbf{Qualitative comparison for rainy2clear (i.e. deraining) on the BDD100K dataset.} The deraining model SAPNet is trained on images from different rain generation methods.}
            \label{fig:vis_derain}
        \end{figure*}

        \begin{figure*}[t]
            \centering


            {\includegraphics[width=2.4cm]{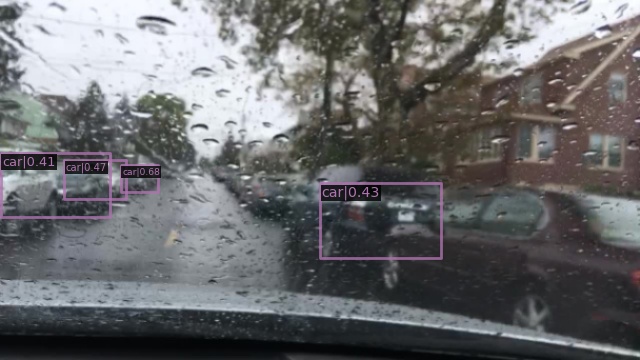}} 
            {\includegraphics[width=2.4cm]{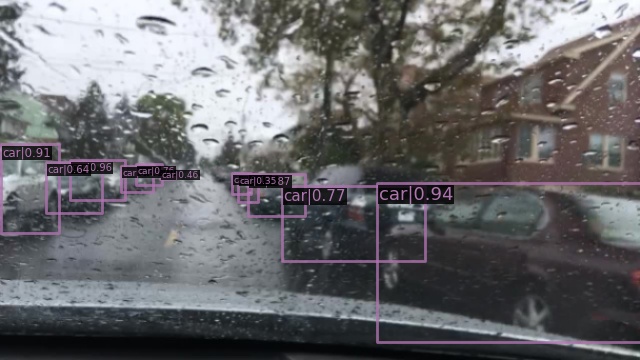}}
           {\includegraphics[width=2.4cm]{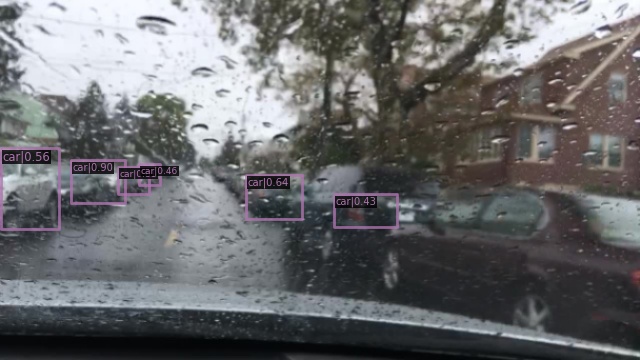}}
             {\includegraphics[width=2.4cm]{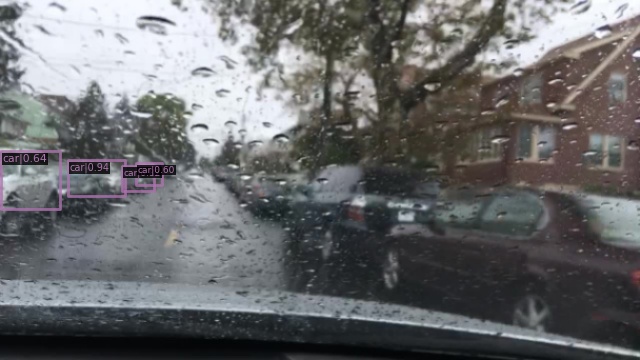}}
           {\includegraphics[width=2.4cm]{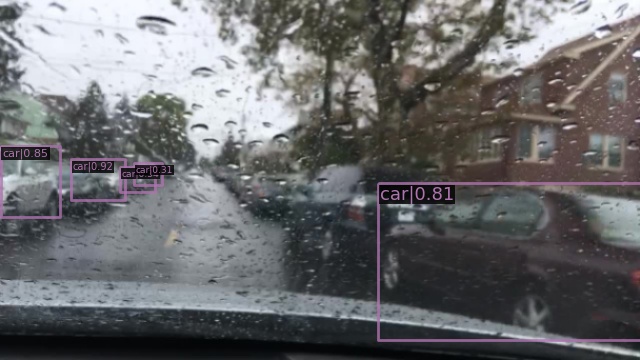}} 
          {\includegraphics[width=2.4cm]{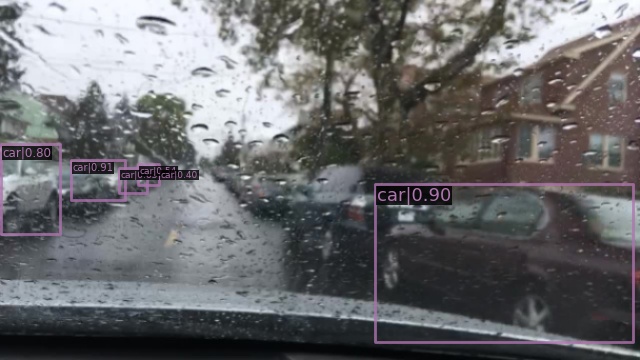}}
           {\includegraphics[width=2.4cm]{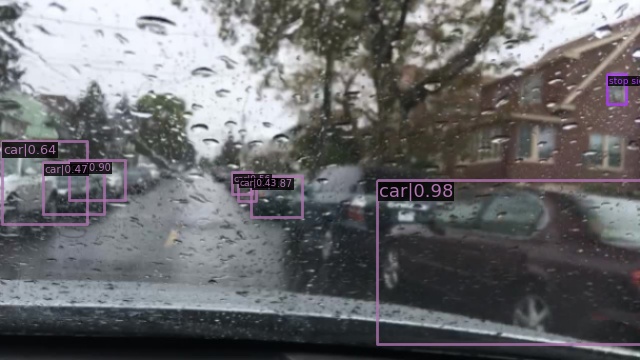}}

            {\includegraphics[width=2.4cm]{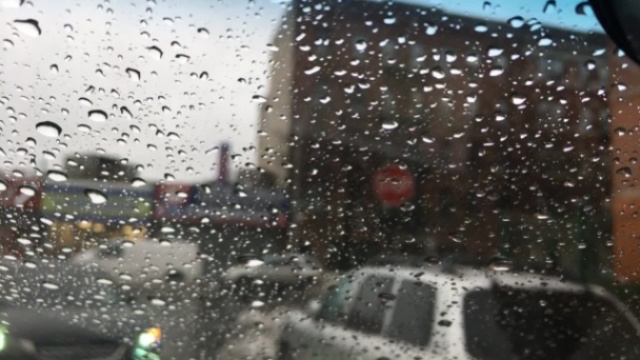}} 
            {\includegraphics[width=2.4cm]{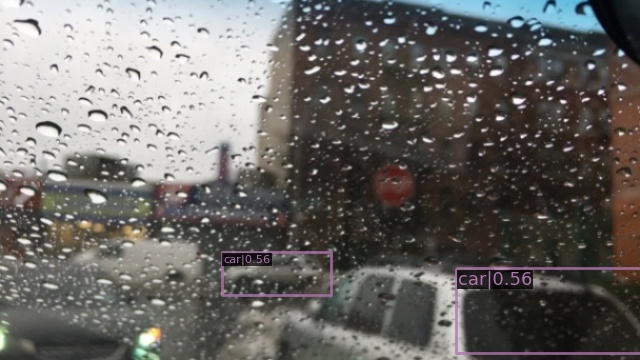}}
           {\includegraphics[width=2.4cm]{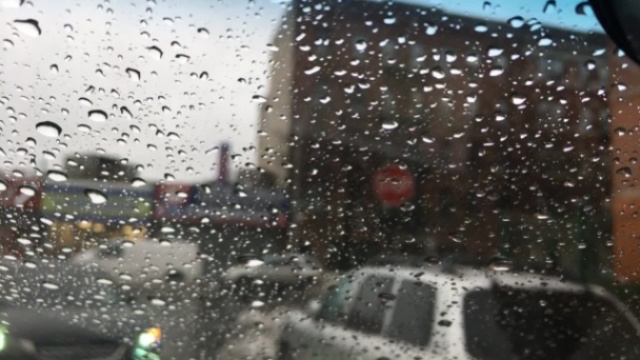}}
             {\includegraphics[width=2.4cm]{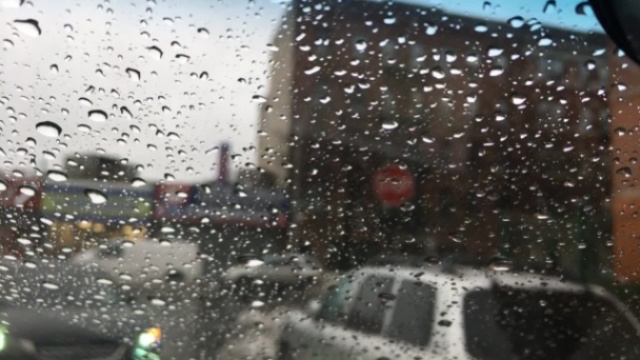}}
           {\includegraphics[width=2.4cm]{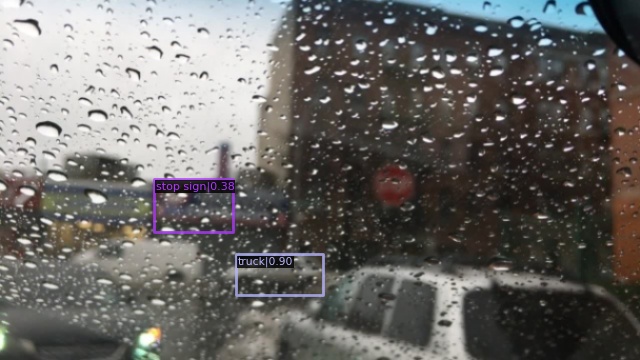}} 
          {\includegraphics[width=2.4cm]{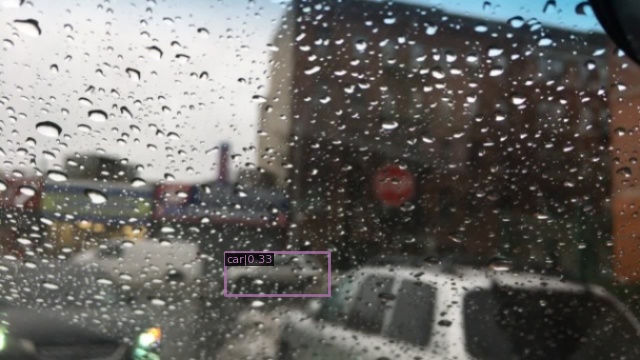}}
           {\includegraphics[width=2.4cm]{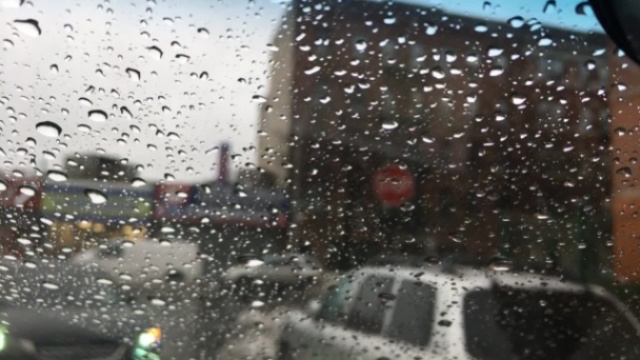}}

        {\includegraphics[width=2.4cm]{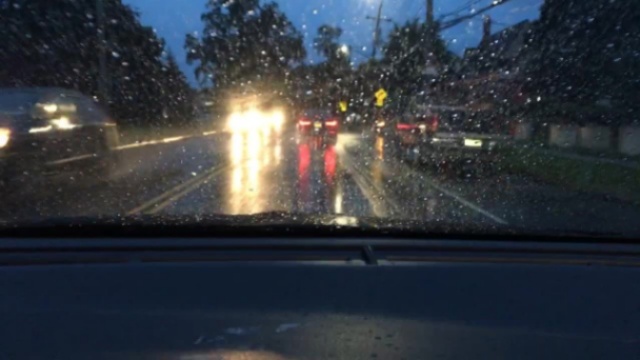}} 
            {\includegraphics[width=2.4cm]{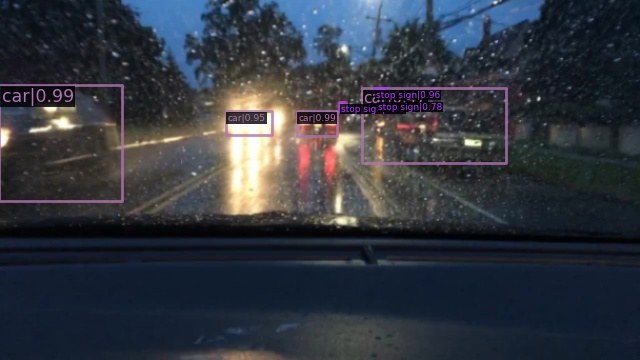}}
           {\includegraphics[width=2.4cm]{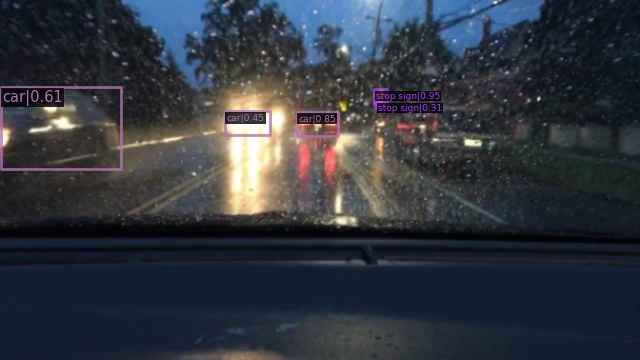}}
             {\includegraphics[width=2.4cm]{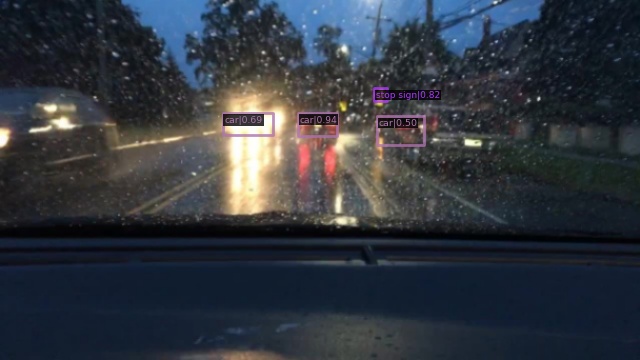}}
           {\includegraphics[width=2.4cm]{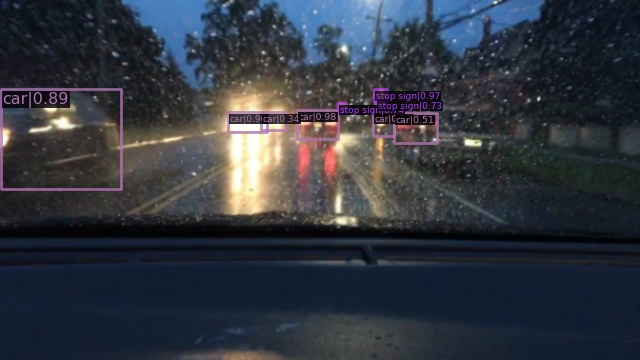}} 
          {\includegraphics[width=2.4cm]{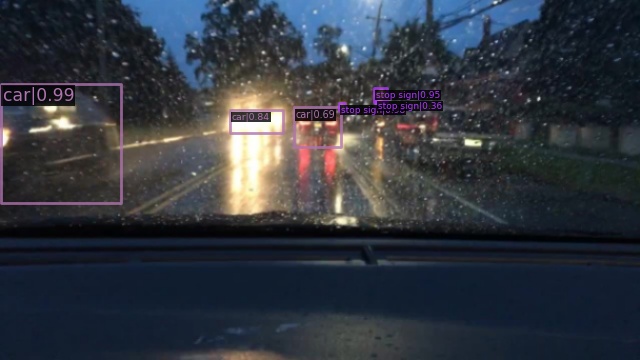}}
           {\includegraphics[width=2.4cm]{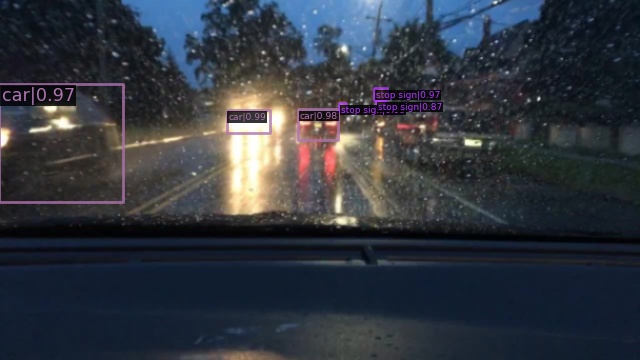}}

            \vspace{-0.15cm}

            \subfigure[Pretrained]{\includegraphics[width=2.4cm]{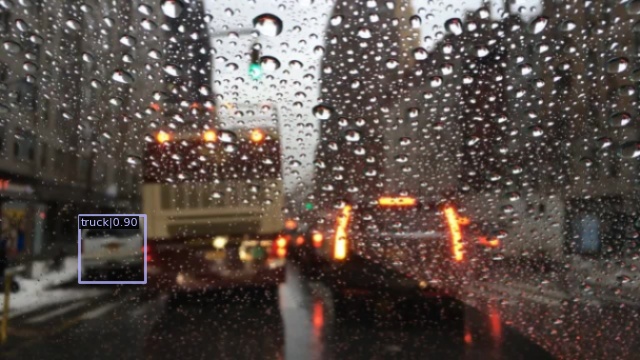}} 
            \subfigure[Ours]{\includegraphics[width=2.4cm]{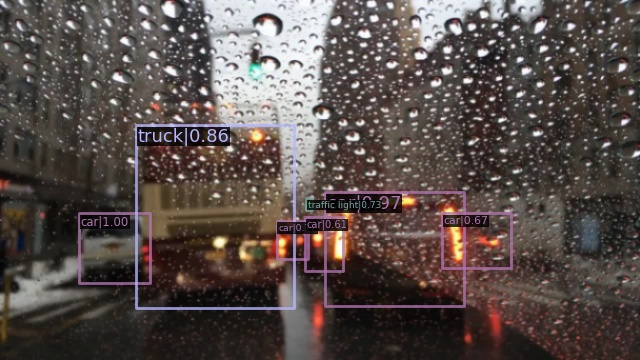}}
            \subfigure[UNIT \cite{liu2017unsupervised}]{\includegraphics[width=2.4cm]{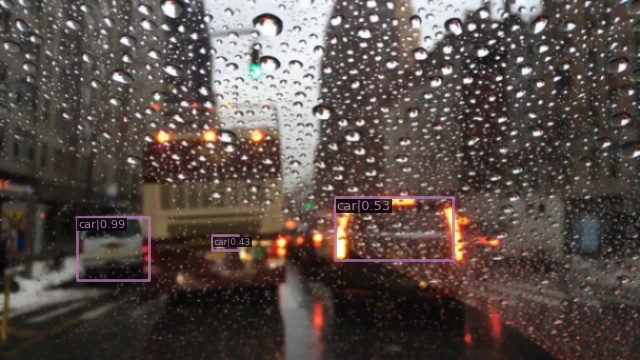}}
             \subfigure[MUNIT \cite{huang2018multimodal}]{\includegraphics[width=2.4cm]{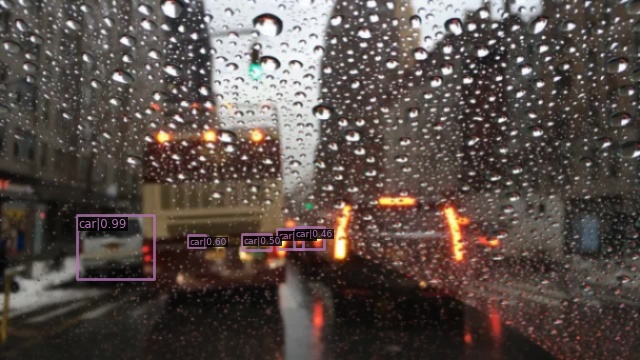}}
            \subfigure[CUT \cite{wu2021contrastive}]{\includegraphics[width=2.4cm]{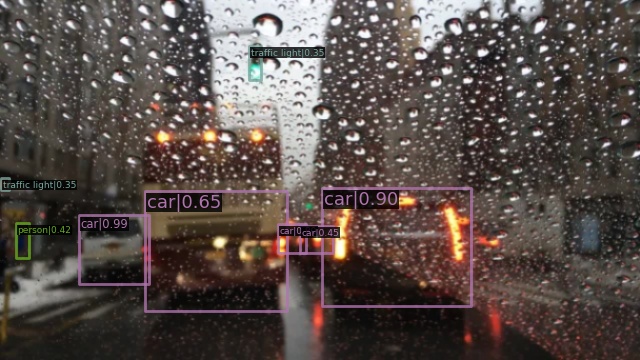}} 
            \subfigure[QS-Attn \cite{hu2022qs}]{\includegraphics[width=2.4cm]{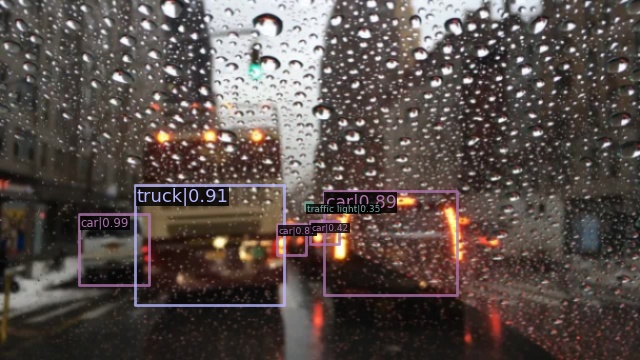}}
            \subfigure[MoNCE \cite{zhan2022modulated}]{\includegraphics[width=2.4cm]{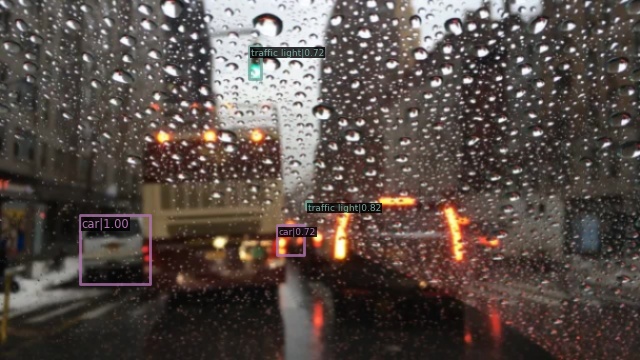}}

            \caption{\textbf{Qualitative comparison for Yolov3 object detection on the BDD100K dataset.} Yolov3 was pretrained on COCO, finetuned on generated rainy images from different rain generation methods, and tested on BDD100K test rainy.}
            \label{fig:vis_detect}
        \end{figure*}

    \subsection{Ablation Studies}

      \textbf{TPS and SeNCE:} We perform an ablation study on the TPS and SeNCE modules. As shown in Fig. \ref{fig:vis_ablation}, the inclusion of TPS \textbf{(M3)} enhances rain generation by mitigating artifacts and distortions. Additionally, the usage of SeNCE \textbf{(M5)} optimizes the contrast and surface water area, leading to more realistic road reflections. Quantitative analysis presented in Tab. \ref{tab:ablation} further demonstrates the efficacy of both TPS and SeNCE in improving rain generation. 

      \textbf{TPS vs. PTL:} We examine the effects of replacing TPS with the related modules Point To Line Distance (PTL). As depicted in Fig. \ref{fig:vis_ablation}, the inclusion of PTL \textbf{(M2)} results in a degradation of rain generation, evident from numerous artifacts and distortions in the background. Conversely, the adoption of TPS leads to high-quality rain generation. Furthermore, Tab. \ref{tab:ablation} validates that PTL yields worse scores, while the proposed TPS achieves significantly better scores.

     \textbf{SeNCE vs. Other NCEs:} We investigate the effectiveness of the proposed SeNCE against other NCEs, including PatchNCE \cite{park2020contrastive} and MoNCE \cite{zhan2022modulated}. Fig. \ref{fig:vis_ablation} shows that SeNCE \textbf{(M5)} leads to high-quality wet surfaces and road reflections, compared with MoNCE \textbf{(M4)} and PatchNCE \textbf{(M1)}. This is confirmed at Tab. \ref{tab:ablation}, where SeNCE leads to generally better scores than MoNCE or PatchNCE. 
    

     \textbf{Semantic Metrics Selection in SeNCE:}  We justify the choice of using mPA compared with widely used mIoU.  Fig. \ref{fig:vis_ablation} shows that TPSeNCE with mPA \textbf{(M7)} surpass TPSeNCE with mIoU \textbf{(M6)} in generating realistic surface water and road reflections. This is agreed by Tab. \ref{tab:ablation}, where mPA leads to much better scores than mIoU.

    \subsection{Experiment Results} 

    \textbf{Rain Generation:}  Tab. \ref{tab:quant_bdd_init} shows  quantitative comparison for rain generation on BDD100K and INIT, where our method attained the best score on all metrics, except for FID on BDD100K. One possible reason is that FID is insensitive to local variations, such as small regions containing artifacts and distortions \cite{gragnaniello2021gan}. Fig. \ref{fig:vis_raingen} showcases a qualitative comparison, revealing that our method generates the most realistic rainy images, with a perfect balance between raindrops, wet surfaces, and road reflections. Besides, our method effectively preserves the content of the clear image without introducing noticeable artifacts or distortions. 
    




    \textbf{Deraining:} Tab. \ref{tab:quant_derain} presents a quantitative comparison of deraining performance on BDD, indicating that our method achieves the highest score on all deraining metrics. Meanwhile, Fig. \ref{fig:vis_derain} provides a qualitative comparison, highlighting that our method outperforms other models in removing raindrops and mists while retaining feature details, restoring color balance, and suppressing noise, blur, and artifacts. 
    
    \textbf{Detection:} Tab. \ref{tab:obj_det_rainmake} presents a quantitative comparison of Yolov3 finetuned on images from different rain generation methods.  The most challenging 100 images in heavy rain or poor lighting were selected for computing mean average precision (mAP) since most rainy images in BDD100K are captured in light rain and good illumination conditions and cannot distinguish the performance of different models. Our proposed method achieved the highest mAP scores across all IoU thresholds and object sizes. Also, as shown in Fig. \ref{fig:vis_detect}, our method outperforms others  in detecting objects under heavy raindrops, with the smallest number of false positives and false negatives across multiple object classes, such as cars, traffic lights, and stop signs.

\vspace{-0.1in}
    \subsection{Extension to Snowy and Night}

\begin{figure}
        \centering


        \includegraphics[width=0.19\columnwidth]{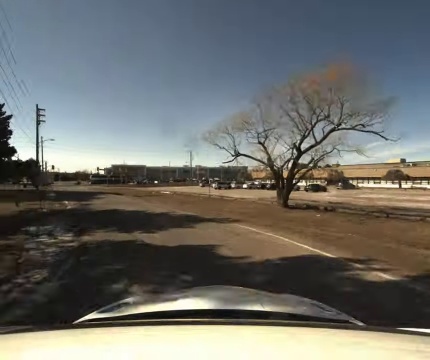}
        \includegraphics[width=0.19\columnwidth]{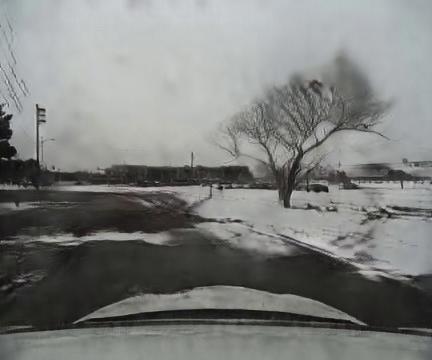}
        \includegraphics[width=0.19\columnwidth]{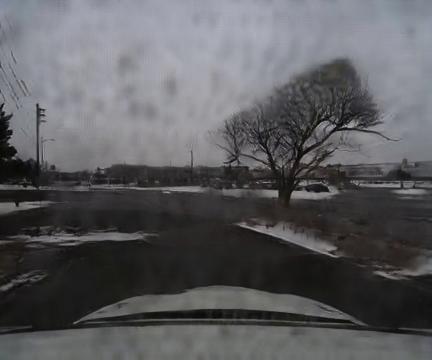}
        \includegraphics[width=0.19\columnwidth]{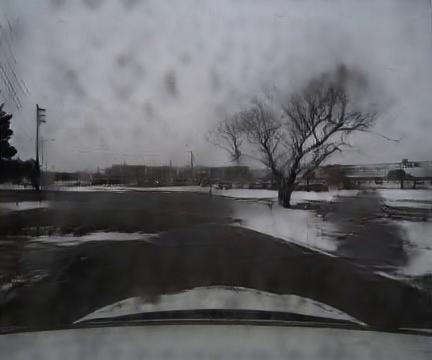}
        \includegraphics[width=0.19\columnwidth]{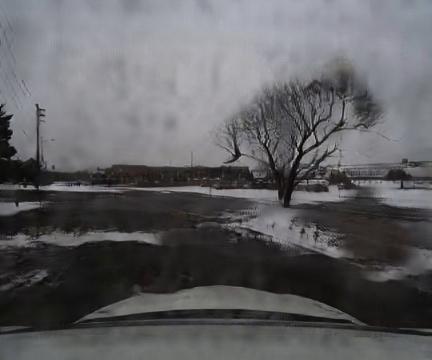}

        \includegraphics[width=0.19\columnwidth]{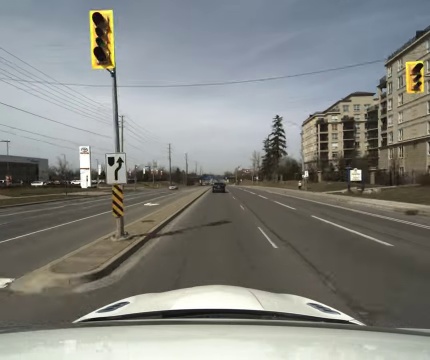}
        \includegraphics[width=0.19\columnwidth]{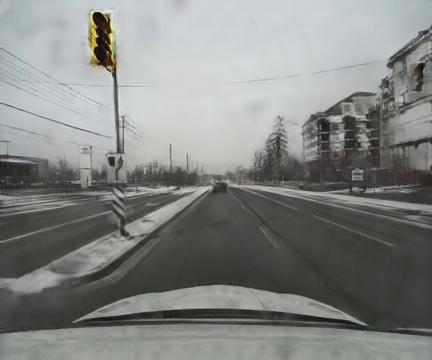}
        \includegraphics[width=0.19\columnwidth]{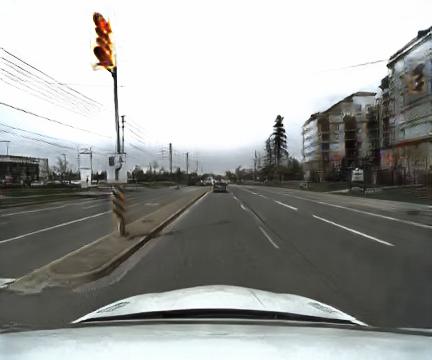}
        \includegraphics[width=0.19\columnwidth]{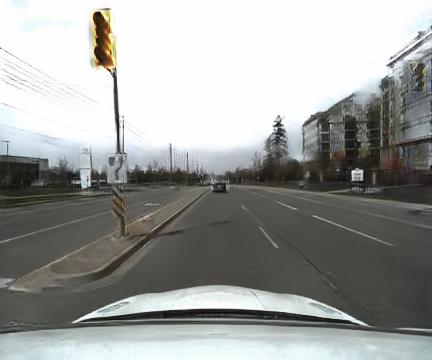}
        \includegraphics[width=0.19\columnwidth]{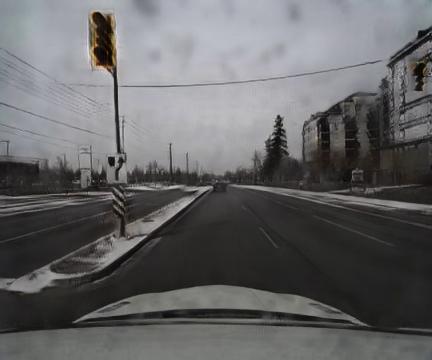}

        \includegraphics[width=0.19\columnwidth]{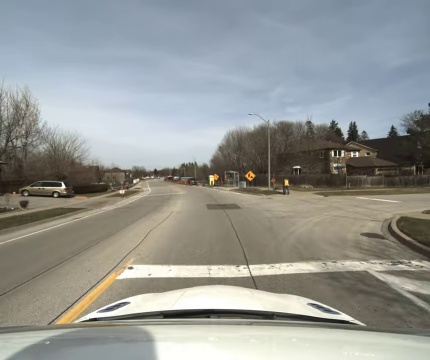}
        \includegraphics[width=0.19\columnwidth]{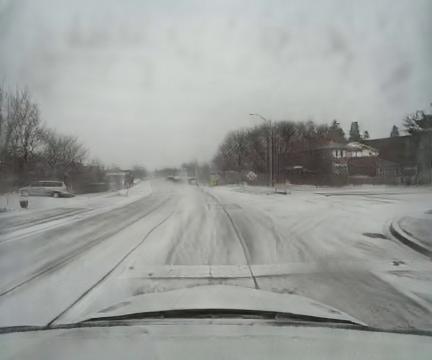}
        \includegraphics[width=0.19\columnwidth]{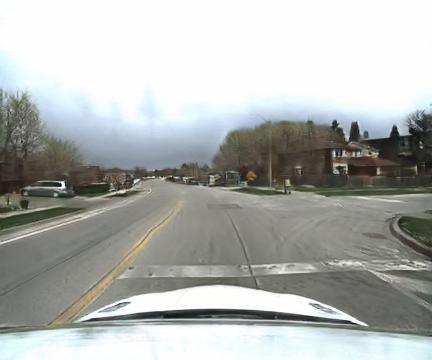}
        \includegraphics[width=0.19\columnwidth]{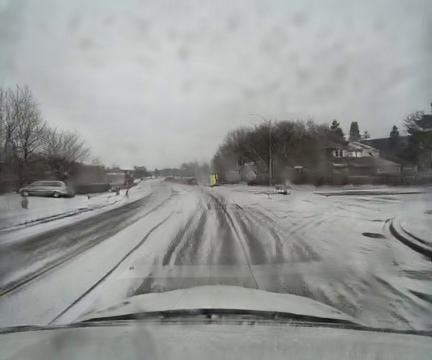}
        \includegraphics[width=0.19\columnwidth]{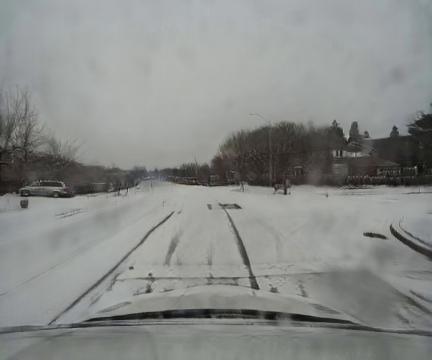}

        \vspace{-0.15cm}

      \subfigure[Clear]{\includegraphics[width=0.19\columnwidth]{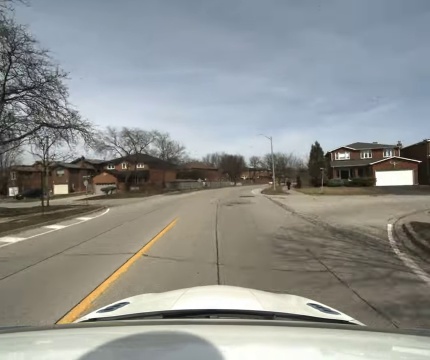}}
        \subfigure[Ours]{\includegraphics[width=0.19\columnwidth]{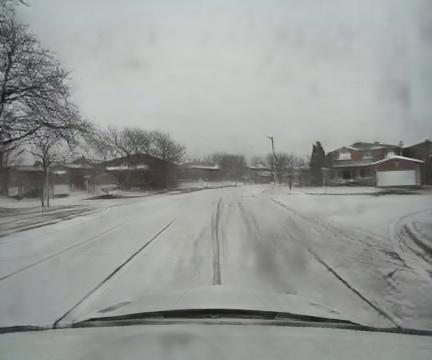}}
        \subfigure[CUT \cite{park2020contrastive}]{\includegraphics[width=0.19\columnwidth]{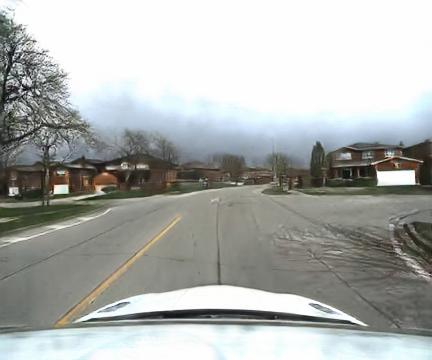}}
        \subfigure[QS-Attn \cite{hu2022qs}]{\includegraphics[width=0.19\columnwidth]{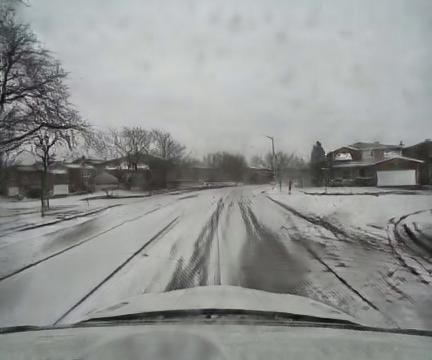}}
        \subfigure[MoNCE \cite{zhan2022modulated}]{\includegraphics[width=0.19\columnwidth]{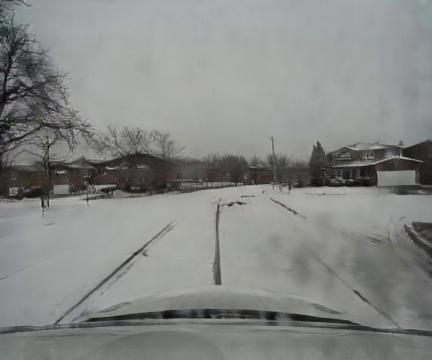}}
        
        \caption{\textbf{Qualitative comparison for clear2snowy translation (i.e. snow generation) on Boreas dataset.}}
        \label{fig:vis_borea_snow}
        \vspace{-0.35cm}
    \end{figure}

     Beyond rain generation, our method works for  \textbf{Clear2Snowy} and \textbf{Day2Night} translation. We show qualitative comparison for snow generation in Fig. \ref{fig:vis_borea_snow}. Our method effectively produces authentic snowy road surfaces with realistic contrast, lighting, reflections, and textures. On the other hand, CUT and QS-Attn produce insufficient snow, while MoNCE's snow lacks realism due to minimal surface reflections and limited contrast variations. Moreover, in Tab. \ref{tab:borea_snow}, our method achieves the best scores for day2night and clear2snowy translation. \footnote{More day2night, clear2snowy translation results are in supplementary. }



    \begin{table}[t]
    \centering
     \setlength\tabcolsep{1.8pt}
    \scriptsize
    \begin{tabular}{c|cccc|cccc}
    \hline
    \multirow{2}{*}{\textbf{Methods}} & \multicolumn{4}{c|}{\textbf{BDD100K Dataset (day $\rightarrow$ night)}}                                                                             & \multicolumn{4}{c}{\textbf{Boreas Dataset (clear $\rightarrow$ snowy)}}                                                                        \\ \cline{2-9} 
                                      & \multicolumn{1}{c|}{\textbf{Content$\uparrow$}} & \multicolumn{1}{c|}{\textbf{Style$\uparrow$}} & \multicolumn{1}{c|}{\textbf{KID$\downarrow$}}    & \textbf{FID$\downarrow$}    & \multicolumn{1}{c|}{\textbf{Content$\uparrow$}} & \multicolumn{1}{c|}{\textbf{Style$\uparrow$}} & \multicolumn{1}{c|}{\textbf{KID$\downarrow$}}    & \textbf{FID$\downarrow$}   \\ \hline
    CUT                               & \multicolumn{1}{c|}{3.10}             & \multicolumn{1}{c|}{3.73}           & \multicolumn{1}{c|}{147.26}          & 16.522          & \multicolumn{1}{c|}{2.80}             & \multicolumn{1}{c|}{1.87}           & \multicolumn{1}{c|}{170.34}          & 36.98          \\ 
    QS-Attn                           & \multicolumn{1}{c|}{3.13}             & \multicolumn{1}{c|}{3.37}           & \multicolumn{1}{c|}{158.83}          & 17.544          & \multicolumn{1}{c|}{3.50}             & \multicolumn{1}{c|}{3.70}           & \multicolumn{1}{c|}{142.96}          & 35.42          \\
    MoNCE                             & \multicolumn{1}{c|}{2.93}             & \multicolumn{1}{c|}{3.30}           & \multicolumn{1}{c|}{142.97}          & 17.003          & \multicolumn{1}{c|}{3.37}             & \multicolumn{1}{c|}{3.57}           & \multicolumn{1}{c|}{158.30}          & 34.62          \\ 
    Ours                           & \multicolumn{1}{c|}{\textbf{3.47}}    & \multicolumn{1}{c|}{\textbf{4.03}}  & \multicolumn{1}{c|}{\textbf{142.10}} & \textbf{15.901} & \multicolumn{1}{c|}{\textbf{4.17}}    & \multicolumn{1}{c|}{\textbf{4.37}}  & \multicolumn{1}{c|}{\textbf{144.40}} & \textbf{33.83} \\ \hline
    \end{tabular}
    \caption{\textbf{Quantitative comparison of day2night translation on BDD100K and clear2snowy on Boreas datasets.}} 
    \label{tab:borea_snow}
    \end{table}


    \subsection{Limitations and Future Work}

    
    Our method is trained on benchmark datasets that compromise mostly of mild rainy images with weak light sources. Consequently, it cannot address extremely heavy rain images effectively. Addressing heavy rain is crucial for improving object detection, given the significant challenges it poses, such as occlusion, reflection, motion blur, low contrast, and noise \cite{yang2020single,zheng2022semantic}. In the future, we intend to gather a large dataset of heavy rain images featuring strong light sources, use physics-based models \cite{li2019heavy,lu2022introvae} for training, and explore joint preprocessing and finetuning \cite{zheng2022low}.


    
    



\section{Conclusion}

This paper presented a novel unpaired image-to-image translation framework that generates highly realistic rainy images with minimal artifacts. Our approach utilizes Triangular Probability Similarity (TPS) and Semantic Noise Contrastive Estimation (SeNCE) to minimize artifacts and distortions and optimize the amount of generated rain. Experiments demonstrate that the proposed method outperforms state-of-the-art in generating realistic rainy images with minimal artifacts, which can benefit image deraining and object detection in rain. Additionally, our method can generate high-quality snowy and night images, highlighting its capability for diverse weather and lighting conditions.

\noindent 
\textbf{Acknowledgements:} This work was supported in part by an NSF CPS Grant CNS-2038612, a DOT RITA Mobility-21 Grant 69A3551747111 and by General Motors Israel.

{\small
\bibliographystyle{ieee_fullname}
\bibliography{egbib}
}

\clearpage


\end{document}